%% file: main.tex
\documentclass[letterpaper]{article} % DO NOT CHANGE THIS
\usepackage[]{aaai25}  % DO NOT CHANGE THIS
\usepackage{times}  % DO NOT CHANGE THIS
\usepackage{helvet}  % DO NOT CHANGE THIS
\usepackage{courier}  % DO NOT CHANGE THIS
\usepackage[hyphens]{url}  % DO NOT CHANGE THIS
\usepackage{graphicx} % DO NOT CHANGE THIS
\usepackage{natbib}  % DO NOT CHANGE THIS AND DO NOT ADD ANY OPTIONS TO IT
\usepackage{caption} % DO NOT CHANGE THIS AND DO NOT ADD ANY OPTIONS TO IT
\usepackage{algorithm}
\usepackage{algorithmic}
\renewenvironment{quote}{%  reduce indentation on quotes
  {\list{}{\leftmargin=0.3in\rightmargin=0.0in}\item[]}%
  {\endlist}}

\usepackage{amsmath} % For the \newtheorem command
\newcounter{example}[section]
\newenvironment{Example}[1][]{\refstepcounter{example}\par\medskip
   \noindent \textbf{Example~\theexample. #1} \rmfamily}{\medskip}

\usepackage{newfloat}
\usepackage{listings}
\DeclareCaptionStyle{ruled}{labelfont=normalfont,labelsep=colon,strut=off} % DO NOT CHANGE THIS
\floatstyle{ruled}
\newfloat{listing}{tb}{lst}{}
\floatname{listing}{Listing}
\nocopyright

\usepackage{xintexpr}  % inserted by agc to do calculations
\usepackage[dvipsnames]{xcolor} 
\usepackage{colortbl} % Added by Rob
\usepackage{booktabs}  % Added by Rob

\newcommand{\systemprompt}[3]{\noindent\textit{\textbf{system:} \textcolor{blue}{#1} \textcolor{black}{#2}  \textcolor{blue}{#3}}}

\newcommand{\lf}{ $\wr$ } %get rid of line breaks, but have option for putting them back in.

\newcommand{\jaccard}{JIx}

\newcommand{\pr}[1]{\textsf{#1}}

\newcommand{\PO}[0]{\pr{PO}}
\newcommand{\TPP}[0]{\pr{TPP}}

\newcommand{\DC}[0]{\pr{DC}}

\newcommand{\EQ}[0]{\pr{EQ}}
\newcommand{\EC}[0]{\pr{EC}}
\newcommand{\NTPP}[0]{\pr{NTPP}}
\newcommand{\NTPPi}[0]{\pr{NTPPi}}
\newcommand{\TPPi}[0]{\pr{TPPi}}

\newcommand{\pct}[2]{#1/#2 (\xintieval[2]{#1 / #2 * 100}\%)}

\newcommand{\claude}{Claude-3.5S}
\newcommand{\gptf}{GPT-4T} % Macros can't have digits in their name
\newcommand{\gptt}{GPT-3.5T}
\newcommand{\gpto}{GPT-4o}
\newcommand{\gemini}{Gemini-1.5P}
\newcommand{\llama}{Llama-3 70B}

\title{Can Large Language Models Reason about the Region Connection Calculus?}
\author {
    Anthony G Cohn\textsuperscript{\rm 1,2} and
    Robert E Blackwell\textsuperscript{\rm 2}
}
\affiliations {
    \textsuperscript{\rm 1}School of Computing, University of Leeds, UK\\
    \textsuperscript{\rm 2}The Alan Turing Institute, UK\\
    a.g.cohn@leeds.ac.uk, rblackwell@turing.ac.uk
}

\begin{document}

\maketitle

\begin{abstract}
Qualitative Spatial Reasoning is a well explored area of Knowledge Representation and Reasoning and has multiple applications ranging from Geographical Information Systems to Robotics and Computer Vision.  Recently, many claims have been made for the reasoning capabilities of Large Language Models (LLMs).  Here, we investigate the extent to which a set of representative LLMs can perform classical qualitative spatial reasoning tasks on the mereotopological Region Connection Calculus, RCC-8.  We conduct three pairs of experiments (reconstruction of composition tables, alignment to human composition preferences, conceptual neighbourhood reconstruction)  using state-of-the-art LLMs; in each pair one experiment uses eponymous relations and one, anonymous relations (to test the extent to which the LLM relies on knowledge about the relation names obtained during training).  All instances are repeated 30 times to measure the stochasticity of the LLMs.  
% The first pair of experiments investigates whether an LLM can reconstruct composition table entries, the second whether an LLM’s preferred relation of a composition coincides with human preferences, and the third whether an LLM can reconstruct the conceptual neighbourhood diagram of RCC-8.  
%The experiments show that, whilst 
%Whilst rather better than chance, none of the LLMs tested performs particularly well. 
No LLM tested performs particularly well, though rather better than chance.
\end{abstract}

% Uncomment the following to link to your code, datasets, an extended version or similar.
%
% \begin{links}
%     \link{Code}{https://aaai.org/example/code}
%     \link{Datasets}{https://aaai.org/example/datasets}
%     \link{Extended version}{https://aaai.org/example/extended-version}
% \end{links}

\section{Introduction }

\begin{figure}[htb]
\centering
\includegraphics[width=0.7\columnwidth]{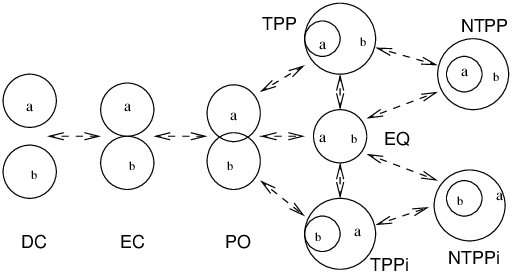} % Reduce the figure size so that it is slightly narrower than the column. Don't use precise values for figure width.This setup will avoid overfull boxes.
\caption{The eight relations of the RCC-8 calculus illustrated in 2D \protect\cite{cohn1997qualitative}:  \DC\ (Disconnected), \EC\ (Externally Connected), \PO\ (Partially Overlapping), \TPP\ (Tangential Proper Part), \NTPP\ (Nontangential Proper Part) and \EQ\ (Equals); \TPPi\  and \NTPPi\ are the inverses of \TPP\  and \NTPP\  respectively since they are asymmetric. }
\label{rcc8-cn-diagram}
\end{figure}

Qualitative Spatial Reasoning (QSR\footnote{We may use QSR as shorthand for both Qualitative Spatial Reasoning and Qualitative Spatial Representation; context should usually make clear which is intended.}) \cite{cohn2008qualitative,chen2015survey,cohn2001qualitative} is a well developed field which is concerned with the representation of qualitative spatial information and reasoning with it.  In natural language, spatial information is usually represented qualitatively  (using prepositions such as \emph{on, in, left of, part of, under, touching, ...)} and many calculi have been developed to represent such information.  There are calculi for mereological relations e.g. RCC-5\footnote{RCC is an acronym for the Region Connection Calculus, which comes in various forms, including RCC-5 and RCC-8}.
%with 5 and 8 jointly exhaustive and pairwise disjoint (JEPD) relations respectively.} 
\cite{DBLP:journals/jair/JonssonD97}, mereotopological relations (such as RCC-8 \cite{randell1992spatial,cohn1997qualitative}, the 9-intersection model \cite{egenhofer-franzosa-9-intersection}), directions (such as OPRA \cite{moratz2006representing}), size \cite{gerevini2002combining} for example, as well as calculi combining two different aspects of spatial information, such as the Rectangle Algebra \cite{guesgen1989spatial,DBLP:conf/aaai/MukerjeeJ90} which can represent both mereotopological information as well as directional.  What is common to all these calculi is that they consist of a set of \emph{jointly exhaustive and pairwise disjoint} (JEPD) \emph{base} relations.  E.g., RCC-8 contains eight JEPD \emph{base} relations, illustrated in 2D in Fig. \ref{rcc8-cn-diagram}.

\emph{Large Language Models} (LLMs) \cite{devlin-etal-2019-bert,brown2020language}, such as GPT-3, LLAMA and GPT-4 
%\cite{roumeliotis2023chatgpt} 
are examples of so called \emph{Foundation Models} \cite{bommasani2021opportunities} which have been trained on very large textual corpora in order to generate text in response to a prompt. This is not the place to survey this burgeoning field, but we note that many claims have been made for the power and apparent intelligent behaviour that  these models can display.  In particular their performance on some benchmarks may lead one to believe that they possess
well developed reasoning capabilities.  So the question arises as to whether
%, at least to some degree, the ability to perform commonsense reasoning \tony{\cite{commonsense-refs}}.  Spatial reasoning is usually regarded as one core aspect of common sense so it is natural to ask whether 
LLMs can perform the reasoning commonly associated with qualitative spatial calculi. This is the question  that we address here.  There are several motivations for this, including whether an LLM could be a replacement for a symbolic spatial reasoner (with the added advantage of being able to pose queries in natural language) and also whether an LLM could be used to construct the reasoning mechanisms for a QSR -- typically these are manually derived (though sometimes  reasoning \cite{randell1992computing,hazarika2005qualitative,bennett1996modal} or machine learning \cite{clementini2024extension} approaches have been taken), and automating the process could ease the production of new QSRs; here we take an existing calculus (RCC-8) as a test vehicle for answering this question. 
%An earlier version of this work (using just a single LLM, ChatGPT) was presented at QR-23, a workshop without archival proceedings~\cite{cohn2023evaluation}.

\section{Related Work}
If we are interested in evaluating the abilities of LLMs to perform spatial reasoning, there are different kinds of spatial reasoning tasks which can be considered.  Relational composition is one of the most studied from a theoretical point of view. R3($x,z$) is the composition of R1($x,y$) and R2($y,z$) if it is implied by the latter two relations.  In general R3($x,z$) is a disjunction of relations. 
%(given a relation R1 between $x$ and $y$, and a relation R2 between $y$ and $z$, what is the relation between $x$ and $z$?). 
A \emph{composition table} (CT) records the results for all combinations of relations in a particular QSR such as RCC.
%an investigation \cite{cohn2023evaluation} into ChatGPT's abilities to compute all RCC compositions
%found reasonable accuracy levels (reduced when  relations are anonymised); however RCC is a
%purely mereotopological calculus with no notion of direction embedded in its semantics. 
A second form of reasoning commonly associated with QSR involves what is called a \emph{conceptual neighbourhood} (CN) -- also called a \emph{continuity network} in \cite{randell1989modelling}. Fig. \ref{rcc8-cn-diagram} illustrates the CN for RCC-8 -- there is an edge between two relations iff one relation can be directly followed by the second one, assuming continuous translation or morphological change. 

Despite the rapidly growing amount of research into LLMs and their capabilities there has been relatively little devoted specifically to spatial reasoning (some exceptions being, e.g. \citet{cohn2024evaluatingabilitylargelanguage,electronics13081532,yamada2024evaluating}), and almost none  exhaustively investigating their abilities to reason about composition in any particular QSR, with the exception of 
\citet{cohn2023evaluation}, 
% an earlier version of this work presented at 
% a closed workshop with no proceedings,
%QR-23,   
which looked at the ability of ChatGPT4 to correctly compute RCC-8's 
%composition table,
CT,
whether it was able to correctly predict human preferred relations in the case of ambiguity; it also investigated whether ChatGPT4 could construct the conceptual neighbourhood diagram of RCC-8. However it only investigated one LLM (ChatGPT-4) and did not investigate the stochasticity of the LLM via repeated experiments.  There are a few works which selectively test compositional reasoning, e.g. \citet{cohn2023dialectical} which investigated a number of spatial reasoning problems include some limited instances of relational composition, but not exhaustively. 
%the already mentioned work
   Other work investigating the spatial reasoning abilities of LLMs typically which revolved around especially constructed  benchmarks such as StepGame \cite{li2024advancing,shi2022stepgame}   can also be regarded as testing compositional reasoning, but not in a methodical or exhaustive manner. StepGame aims to test an LLM's ability to correctly determine the qualitative direction relationship between two objects, given a set direction relations between a larger set of objects, and between 1 and 10 reasoning steps are required to correctly determine the result. 
Not surprisingly, performance deteriorates as the required number of steps increases. Performance increases markedly when the LLM is used to translate from the English specification to a logical representation and symbolic reasoning is used to compute the relationship. The SpartQA dataset \cite{mirzaee-etal-2021-spartqa} is also focused on assessing spatial reasoning, but does not test composition or conceptual neighbourhoods.  The bAbI dataset \cite{weston2016towards}  also has some tasks which test spatial reasoning and compositional reasoning, in particular about directions to a limited extent. Other work  has investigated whether LLMs can acquire an understanding of a spatial environment from a turn-by-turn description of a route, with landmarks named at each turn; whilst the LLMs did perform reasonably well, the experiment did not involve any mereotopological relations, only left/right and
up/down~\cite{yamada2024evaluating}.

%In earlier work \cite{cohn2023dialectical} we use extended dialogues with an LLM to try to map the 
%boundaries of spatial commonsense in some LLMs, addressing a variety spatial challenges, and
%examining not only the response given but also the explanation/justification of the response, but did not specifically focus on existing QSRs, though some questions were asked which do correspond to particular 
%reasoning steps in an existing QSR.  Here we focus on one specific QSR and ask the question as to what extent an LLM %can perform reasoning in that calculus, and conduct a more exhaustive evaluation, but looking 
%at the ability to perform compositions between relations and also to reason about the 
%conceptual neighbourhood diagram of the calculus. 
Weaknesses in the reasoning powers of LLMs have previously been noted (e.g. \citet{cai2023human}) so one might not expect LLMs to perform well in this regard. But on the other hand, there are a large number of papers about QSR in the literature and these are likely to have formed part of the  training corpus of an LLM, and thus might facilitate correctly responding to prompts --  though the information concerning the actual reasoning steps are often given in tables (in particular 
%\emph{composition tables}
\emph{CT}s -- see below) and thus might be hard for LLM training procedures to process effectively.

\section{Experimental Design}
There are now many LLMs in the literature, with new ones released frequently.  Some of these are open source and are explicit about the training corpus; others are closed and give no specific information about the training, or the precise corpus, such as the GPT family of LLMs.  
Previous work suggests that models with less than about 40B parameters perform poorly at reasoning (e.g. \citet{leyton-brown-slides}). We therefore favour larger models, testing those  in Table~\ref{tab:models}. 
 
LLMs are stochastic in nature and show considerable variability in their answers. Vendors provide various API options (e.g., \emph{seed}, \emph{temperature}, and \emph{top\_p}) to try to make sampling more consistent. However, no settings that we have yet tried (including setting \emph{temperature} to 0) result in fully deterministic answers (e.g. see Appendix Fig. \ref{rcc8-variability}).

We therefore accept all model defaults and repeat each chat completion multiple times. 
We choose $n=30$ repeats being the sample size at which, according to the Central Limit Theorem, the sampling distribution of a mean approximates normality well enough for practical purposes. 
To measure accuracy, we use the mean plus or minus the 95\% confidence interval of the mean of answer scores.
In all of the experiments each question was posed in a separate conversation. 
We switched off the guard rails for models hosted on Azure OpenAI. 
We set \emph{max\_tokens} to 512 in the Anthropic API, it being a required parameter.

Although there are many different prompt engineering strategies \citep[see][]{schulhoff2024prompt}, our purpose here is to test, rather than optimise, model performance. We therefore prefer simple, natural prompting. Our only concession to prompt engineering is to add
\textit{Answer the question and provide the final answer in the form: \texttt{"}\#\#\# Answer:\texttt{"}} to facilitate pattern matching and automation of answer assessment using regular expressions.

In the experiments below, since multiple relations may be possible as an answer, we use the Jaccard Index (\jaccard, \citet{jaccard1901etude}) to compute the accuracy of the predicted response compared to the expected, ground truth, answer. The \jaccard\  is calculated by counting the size of the intersection of the predicted set of relations and the ground truth set,  divided that by the number of relations in the union of the two sets. When only a single relation is given and the expected answer is just a single relation, this reduces to a binary 1 or 0 measure of accuracy. 

Details of the computing infrastructure used to run the experiments can be found in the appendix.  The results of all runs, and the programs used to do the experiments are in the uploaded  supplementary material. 

\begin{table*}[htb]
\footnotesize
\centering
%\begin{tabular}{|m{2.5cm}|m{4cm}|m{1.5cm}|m{1cm}|m{1.5cm}|}
\begin{tabular}{lllllrr}
\toprule
\textbf{Vendor} &  \textbf{Model} & \textbf{Abbreviation}&\textbf{Released} & \textbf{API} & \textbf{Number of  Parameters} & \textbf{Context} \\ \hline
%Anthropic Claude& claude-3-opus-20240229 & Feb 2024 & 137B & 200,000  \\ 
%Google Vertex & gemini-10-pro & Dec 2023 & 1.6T &  32,000   \\ 

%OpenAI & gpt-3.5-turbo-0613 \cite{brown2020language}  & Jun 2023 & 175B & 4,096   \\ 
%  & gpt-35-turbo-1106 & Nov 2023 & 175B & 16,385  \\ 
%OpenAI & gpt-3.5-turbo-0125 & Jan 2024 & 175B & 16,385  \\ 
%& gpt-4-0613 \cite{openai2024gpt4} & Jun 2023 & 1.76T & 8,192   \\ 
%  & gpt-4-turbo-2024-04-09 & Apr 2024 & 1.76T & 128,000   \\ \hline
OpenAI & GPT-3.5 Turbo 0125 &  \gptt & Jan 2024 & Azure OpenAI & Undisclosed & 16K  \\ 
Meta &  Llama 3 70B Instruct &\llama &Apr 2024 & Azure AI & 70B &8K \\
Google &Gemini 1.5 Pro  preview-0409 & \gemini& Apr 2024 & Google Vertex & Undisclosed, est. $\gg$3.5T & 128K   \\ 
OpenAI & GPT-4 Turbo 2024-04-09 &\gptf & Apr 2024 & Azure OpenAI & Undisclosed& 128K  \\ 
OpenAI & GPT-4o  2024-05-13 & \gpto & May 2024 & Azure OpenAI & Undisclosed & 128K  \\ 
Athropic &  Claude 3.5 Sonnet  20240620 &\claude & Jun 2024 & Anthropic & Undisclosed & 200K \\
%Mistral AI & \mistral & Sep 2023 & Ollama, Mac M2 & 7B & 8K \\
%Meta &   \sllama & Apr 2024 & Ollama, Mac M2 &7B & 8K \\
%\hline
\bottomrule
\end{tabular}
\caption{LLMs tested.
%in our experiments.
%Size is the number of parameters. 
%Params is the number of parameters where known.
Context is the context window size in tokens.
Models will be referred to by their abbreviation henceforth.}
\label{tab:models}
\end{table*}

\section{Compositional Reasoning}
The most researched form of QSR reasoning  
%Qualitative Spatial Relations 
is that of composition: i.e. given two relations R1($x,y$), and R2($y,z$), then which relations are possible between $x$ and $z$? In general, 
%more than one base relation is possible, so 
the answer may be a disjunction of relations, as can be seen in the RCC-8 CT in Fig. \ref{ct-jaccard} (which also uses colouring to show the results of Experiment 1 below).  Note that this table omits the rows and columns for the \EQ\  relation (we do not include compositions involving \EQ\ in our experiment since they should be trivial).
%\ref{rcc8comptable-exp1}. 
Given the ubiquity of RCC-8 in the QSR literature, in this paper we focus on the abilities of an LLM to reason with RCC-8.  

% \begin{figure*}[htb]
% \centering
% \includegraphics[scale=0.5]{rcc8-composition-table.png}
% %\includegraphics[width=0.8\textwidth,height=0.5in]{rcc8-composition-table.png} % Reduce the figure size so that it is slightly narrower than the column. Don't use precise values for figure width.This setup will avoid overfull boxes.
% \caption{The RCC-8 Composition Table (CT) \protect\cite{cohn1997qualitative}. = means EQ and * means ALL}
% \label{rcc8comptable}
% \end{figure*}

\subsection{Experiment 1 : Compositional Reasoning in RCC-8}
\begin{figure}[ht]
\centering
\includegraphics[width=0.99\columnwidth]{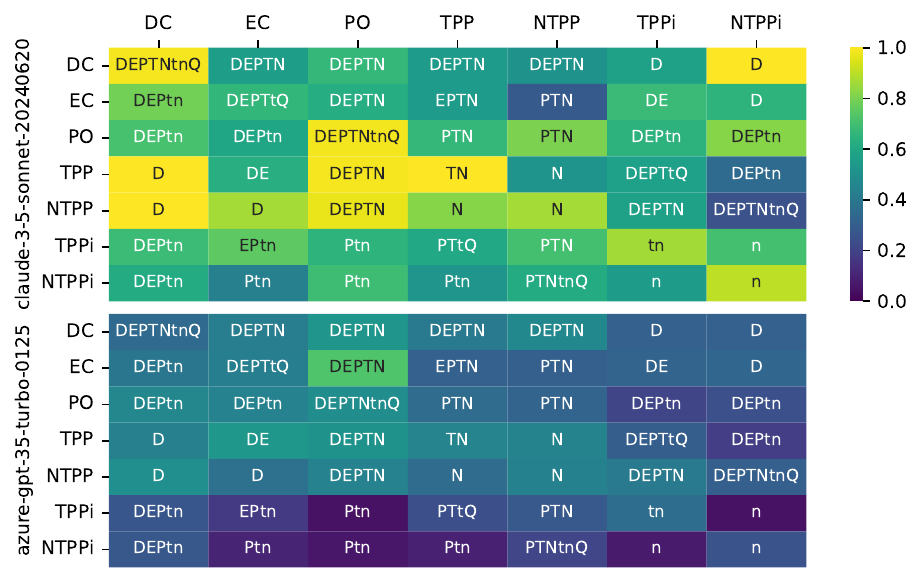} % Reduce the figure size so that it is slightly narrower than the column. Don't use precise values for figure width.This setup will avoid overfull boxes.
\caption{RCC8 CT shaded by mean Jaccard Index (n=30 repeats) for the best performing model, \claude\ and the worst performing model \gptt. The entry in each cell uses the following coding: {\sf D (DC), E (EC), P (PO), T (TPP), N (NTPP), t (TPPi), n (NTPPi), Q (EQ).}   The full results are in the appendix, Table \ref{rcc8-composition-table}.
%Labels show ground truth answers encoded using the same notation as Table \ref{rcc8-composition-table}.
 }
\label{ct-jaccard}
\end{figure}

In this experiment the standard names of the eight relations were given with a sentence stating that the relations were part of the RCC-8 calculus, as well as English language descriptions of their semantics. Each prompt concluded with a question asking for the relations involved in one particular composition. 
We generated 7x7=49 prompt questions based on the RCC-8 CT (Fig. 
%\ref{rcc8comptable}
\ref{ct-jaccard}).
%(compositions involving \EQ\  were omitted since they should be trivial).
For example\footnote{To save space, in the {\bf system:} prompt, the text in blue is standard across all the system prompts in the experiments in this paper and will not be repeated subsequently, only the text in black, the missing text indicated by ellipsis. Also in the {\bf assistant:} responses, we remove all the blank lines generated by the LLM, replacing each blank line by a $\wr$ symbol.}:

\begin{quote}
%\textbf{system:}
\systemprompt{You are a helpful assistant. Consider the following set of eight pairwise disjoint and mutually exhaustive binary spatial relations. These relations form part of the well known RCC-8 qualitative spatial reasoning calculus. DC(x,y) means that x and y are disconnected and share no spatial parts. EC(x,y) means that x and y touch at a boundary but do not share any interior parts. PO(x,y) means that x and y share a spatial part, but neither is part of the other.  TPP(x,y) means that x is part of y and touches y’s boundary. NTPP(x,y) means that x is part of y but does not touch y’s boundary.  TPPi(x,y) is the same as TPP(y,x).  NTPPi(x,y) is the same as NTPP(y,x).  Finally, EQ(x,y) means that x and y are coincident. I will now ask you a question about these relations.}{There may be more than one possible relation, in which case name all of the possible answers.}{ Answer the question and provide the final answer in the form: "\#\#\# Answer:"}

\noindent\textbf{user:} 
\textit{If DC(x,y) and DC(y,z) then what are the possible relationships between x and z?}
\end{quote}

\begin{figure}[htb]
\centering
\includegraphics[width=0.99\columnwidth]{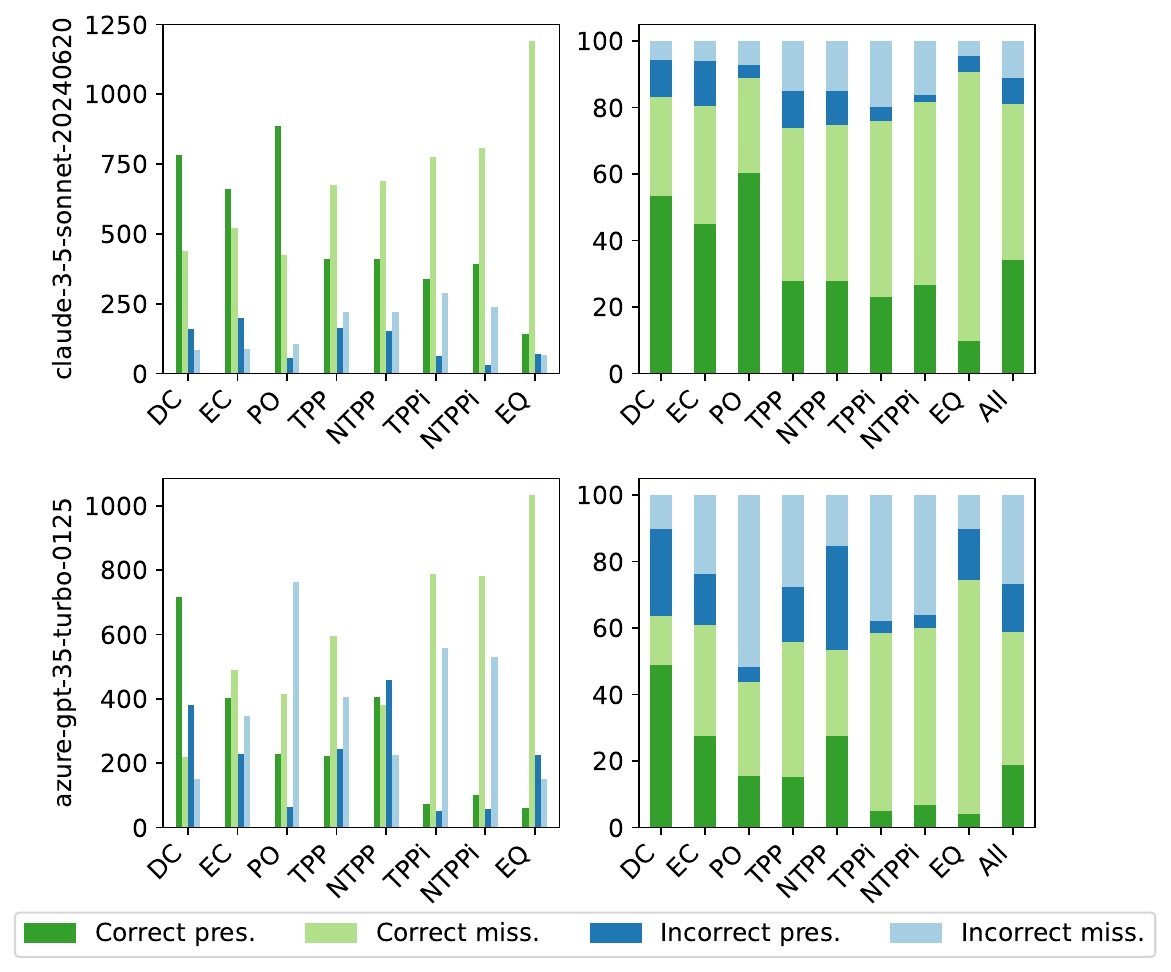} % Reduce the figure size so that it is slightly narrower than the column. Don't use precise values for figure width.This setup will avoid overfull boxes.
\caption{Relation statistics for the CT for RCC-8. The left hand chart shows the absolute number of relations, and the right hand the relative percentage for each relation across all thirty repeats. \emph{All} is the aggregate of all relations.
 }
\label{ct-relation-statistics}
\end{figure}

 The best performing model for experiment 1   is \claude,
 %\ ($0.69 \pm 0.008$), 
 followed by \gemini;
 %\ ($0.51 \pm 0.013$); 
 the worst performing  is \gptt\ 
 %\ ($0.33 \pm 0.009$) 
 (Table \ref{overall-summary},  column ``1epon'').  
 All models performed better than guessing ($> \approx 0.31$).  More detailed statistics can be found in Fig. \ref{ct-relation-statistics}.
%\claude's first answer was as follows:
\claude's first answer (Appendix Example \ref{answer-1}) is correct, well justified, and scores 1.0. 
%This answer is correct, well justified, and scores 1.0. 
However, repeat 13 from \claude\ was:

\begin{quote}
\noindent\textbf{assistant:} 
{\itshape
To answer this question, we need to consider all possible relationships between x and z given that x is disconnected from y, and y is disconnected from z.
\lf
\lf
\lf
Given:
\lf
- DC(x,y): x and y are disconnected and share no spatial parts
\lf
- DC(y,z): y and z are disconnected and share no spatial parts
\lf
\lf
\lf
Possible relationships between x and z:
\lf
\lf
\lf
1. DC(x,z): x and z could be disconnected from each other, just as they are both disconnected from y.
\lf
\lf
\lf
2. EC(x,z): x and z could be externally connected, touching at their boundaries without sharing interior parts.
\lf
\lf
\lf
3. PO(x,z): x and z could partially overlap, sharing some spatial parts but not being fully contained within each other.
\lf
\lf
\lf
4. TPP(x,z) or TPPi(z,x): x could be a tangential proper part of z, or vice versa.
\lf
\lf
\lf
5. NTPP(x,z) or NTPPi(z,x): x could be a non-tangential proper part of z, or vice versa.
\lf
\lf
\lf
6. EQ(x,z): x and z could be equal or coincident.
\lf
\lf
\lf
In this scenario, all possible RCC-8 relations between x and z are possible because the given information about their relationships to y does not constrain their relationship to each other in any way.
\lf
\lf
\lf
\#\#\# Answer:
\lf
DC(x,z), EC(x,z), PO(x,z), TPP(x,z), NTPP(x,z), TPPi(z,x), NTPPi(z,x), EQ(x,z)
}
\end{quote}

This answer is only partially correct (\TPPi(z,x) is equivalent to \TPP(x,z) but should be \TPPi(x,z) and \NTPPi(z,x) is equivalent to \NTPP(x,z) but should be \NTPPi(x,z)) so this answer scores 0.75.

%Results are summarised in Table \ref{rcc8-composition-table} and Figure \ref{rcc8-comp-relation-stats}.

% There are 49 questions * 30 repetition = 1470

Although we asked for possible relations between x and z, models sometimes reply with relationships between z and x (e.g., \claude\ gave \pct{288}{1470} answers that included a relation R(z,x)). Generously, we therefore interpret TPPi(z,x) as TPP(x,z) and NTPPi(z,x) as NTPP(x,z) and vice versa. We also allow the symmetric relationships \DC(z,x),\ \EC(z,x),\ \PO(z,x) and \EQ(z,x).

%\claude\ was the best performing model with a mean \jaccard\  of ($0.69 \pm 0.008$), and \gptt\ had the lowest \jaccard\   ($0.33 \pm 0.009$, Table \ref{overall-summary} column 1epon).

\claude\ got \pct{513}{1470} answers completely correct, \gemini\ \pct{307}{1470} and \gptt\ only \pct{47}{1470} (the lowest).

\claude\ predicted \DC\ most frequently \pct{944}{4937} and \EQ\ least frequently \pct{212}{4937}. \claude\ gave \pct{14}{4937} invalid relations (i.e. involving y, rather than only x and z), \EC(x,y), \EC(y,x), \PO(y,z) and \PO(x,y). \gptt\ predicted \DC\ most frequently \pct{1100}{3925} and \TPPi\ least frequently \pct{124}{3925}. \gptt\ gave \pct{3}{3925} invalid relations, \EC(y,z), \NTPP(x,y), and \TPP(y,x). Although few, these invalid relations
suggest a fundamental lack of understanding of what composition is.

\claude\ predicted \TPP\ more frequently
\pct{574}{4937} than \TPPi\ \pct{404}{4937} and \NTPP\ more frequently \pct{563}{4937} than \NTPPi\ \pct{424}{4937}. Similarly,
 \gptt\ predicted \TPP\ more frequently
\pct{466}{3925} than \TPPi\ \pct{124}{3925} and \NTPP\ more frequently \pct{865}{3925} than \NTPPi\ \pct{157}{3925}. These statistics suggest a bias away from the
inverse in these asymmetrical relations. Indeed these four relations all tend to have fewer correct answers in comparison to other relations (Fig. \ref{ct-relation-statistics}).

Whilst the original CT in Fig. \ref{rcc8-cn-diagram} has a certain symmetry around the leading diagonal (after substituting the inverse relations for \TPP\ and \NTPPi\  where appropriate and switching to the relevant rows -- e.g. (\TPP{\tiny$^\circ$}\EC) has the same predicted relations as (\EC{\tiny$^\circ$}\TPPi)), the LLM produced ones are not, and this is in no small part due to their reluctance to include the inverse relations. 

%The subsequent questions  followed the same format as in Experiment 1, but with the standard relation names.
%The results of the 7*7=49  prompts issued (compositions involving \EQ\  were omitted since they should be trivial) are summarised in  Fig. \ref{rcc8-composition-table}.  
% There is a total of 85 relations correctly predicted (those in black),  197 relations which are correctly not predicted (those not in the table). Thus there are a total of 282 correctly predicted (both present and missing) relations, i.e. 71.94\% of the total possible (7*7*8 = 392). 11 of the cells (out of 49) are completely correct. This is clearly much better than chance but far from a correct CT.  Fig. \ref{rcc8-comp-relation-stats} shows the absolute number and relative percentages of the 8 relations in the proportion that they are correctly or incorrectly predicted. As can be seen, \DC\  is never missed as a correct prediction, and \PO\  is never incorrectly predicted as present. \DC\  has the highest accuracy which is perhaps not surprising since \DC\  might be regarded as intuitively easier to reason about (and this fits with the predictions about human preferred models in the section below). Including both correctly predicted and correctly missing relationships, all eight relations have broadly similar accuracies.

\begin{figure}[ht]
\centering
\includegraphics[width=0.99\columnwidth]{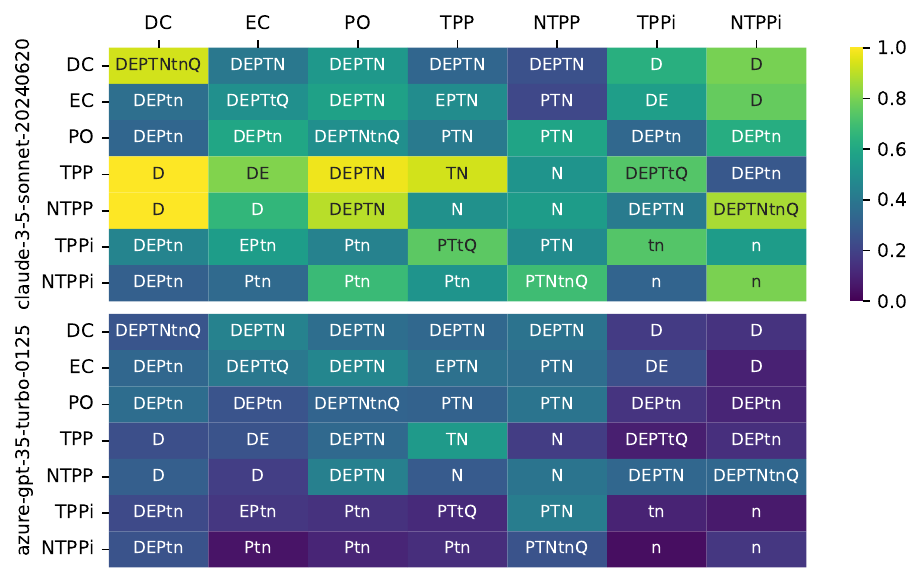} % Reduce the figure size so that it is slightly narrower than the column. Don't use precise values for figure width.This setup will avoid overfull boxes.
\caption{Same as for Figure \ref{ct-jaccard} but  for anonymous.
%Labels show ground truth answers encoded using the same notation as Table \ref{rcc8-composition-table}.
 }
\label{xct-jaccard}
\end{figure}

\rm
In order to test whether the result was influenced by prior knowledge of RCC-8 gained during training, we also performed the same experiment, but with all the relation names anonymised\footnote{%To anonymise the relations, we 
We asked Chat-GPT to make up eight words 
%that are 
not in the dictionary and then mapped  \DC\  to \textit{fablon},  \EC\  to \textit{narkil}, \PO\  to \textit{quonty}, \NTPPi\  to \textit{piflox}, \TPPi\  to \textit{dregly}, \NTPP\ to \textit{lufrex}, \TPP\  to \textit{zorpin}, and \EQ\  to \textit{womfer}. After running the experiments we discovered that fablon is actually a kind of sticky-backed fabric.
%, but this is still a much more robust anonymisation than that used in \cite{cohn2023evaluation} which simply prepended an X to each relation name. 
} 
to disguise the connection to RCC-8. The prompt was the same as above except for the change of relation names and the omission of the third sentence. 
All models showed lower performance with anonymised relations (see Fig. \ref{xct-jaccard}), and the \gptt\ performance was worse than guessing 
%($0.25 \pm 0.011$ $<~\approx0.31$,
(Table \ref{overall-summary}, column 1anon). These results support the view that models have difficulty generalising the information contained in the 
%RCC-8 
literature and cannot reason well about spatial information from first principles.

% Results are summarised in Table \ref{xrcc8-composition-table} and Figure \ref{xrcc8-comp-relation-stats}.

% \claude\ got \pct{356}{1470} answers completely correct, \gpto\ \pct{328}{1470}, \gptf\ \pct{312}{1470}, and \gemini\ \pct{277}{1470}. Despite this, \gemini\ had a higher Jaccard index that \gpto\, getting more answers partially correct.

% Overall, \claude\ had the highest Jaccard Index (0.58 $\pm$ 0.012, Table \ref{overall-summary}).

% The results are given in Fig. \ref{xrcc8-composition-table} while Fig. \ref{xrcc8-comp-relation-stats}  shows the absolute number and relative percentages of the 8 relations in the proportion that they are correctly or incorrectly predicted. 

%As can be seen, \DC\  again is never missed as a correct prediction, and \EC\  is only missed twice; again \PO\  is never incorrectly predicted as present. As before, \DC, \EC, and  \PO   have the highest accuracies, along with \EQ, but \EQ\ is never predicted as present correctly, only incorrectly.  The overall average of correctly predicted relations (present and missing) drops from 71.94\% in the non-anonymous case above  to 67.09\% so there is some loss of performance though whether is due to the anonymisation of the relations or the stochasticity of \chat\  is not clear.

\subsection{Experiment 2 : Preferred Compositions in RCC-8}

As noted above, in general a composition of two relations will yield more than one possible base relation, but it turns out that humans tend to have a ``preferred'' relation.  For example, Ragni et al's \shortcite{ragni2007cross} report on experiments performed on native German 
%speakers 
and native Mongolian speakers (20 of each) for RCC-8. 
%In their experiments the 
The
relations were described, but the 
%human 
subjects were not allowed to draw possible configurations, so the setting is essentially equivalent to an LLM setting.

Given that humans may struggle to see all the possible relations\footnote{The fact that some humans may struggle to compute the CT does not stop  it being a valid question to see if an LLM can determine the correct entries. Indeed, given that it has already been remarked upon that determining the correct CT is challenging \cite{randell1992computing}, if LLMs could automatically compute CTs for new sets of relations, this would be very helpful.}, determining whether there is agreement about the most preferred is a good question to ask.  It turns out that there is good agreement in general across and within the two cultures, with the the percentage of people agreeing with the same preferred relation ranging from 30\% to 87.5\% (a random choice would yield 12.5\% on average since there are eight relations to choose from). (They did not query cases where the composition yields a unique relation, nor did they   consider \EQ\ as one of the two relations as this should be a trivial task.) This agreement is perhaps surprising since the two languages are linguistically very different. Ragni et al \shortcite{ragni2007cross} do report some differences though -- for example although both language speakers preferred \DC\  whenever it was consistent, Mongolians preferred \PO\  over \NTPPi\  whereas for Germans the converse was true. Both cultures only chose \EQ\  when composing a relation with its inverse (e.g. \TPP\  with \TPPi). 

We found an apparent error when transcribing the preferred relations
from Table 2 in Ragni et al \shortcite{ragni2007cross} . The preferred
relation for \NTPPi\ followed by \TPP\ is given as \PO\, but the mean score
is given as 45\% rather than 35\% and so we use the \NTPPi\ (40\%) relation
instead.

The theory of \emph{preferred mental models} \cite{knauff1995preferred} states that people construct the simplest (computationally cheapest) model consistent with the premises. Their experiments showed that humans prefer models with the smallest overlapping complexity which explains the preference for \DC\  noted above.

Given the difficulty reported in Experiment 1 that LLMs have in  correctly inferring all possible relations in a composition, asking this more specific query seems a natural task to try; this gives some measure of whether the cognitive preferences of humans correspond at all to LLM choices.
%made by an LLM.  

%The system prompt given to the LLMs was similar to the initial prompt above when querying about composition. 

% Some responses contained a clear preference with a good argument, for example, in the following response from \claude\ where \DC\ is indeed the preference reported by \cite{ragni2007cross}. 

\begin{figure}[ht]
\centering
\includegraphics[width=0.99\columnwidth]{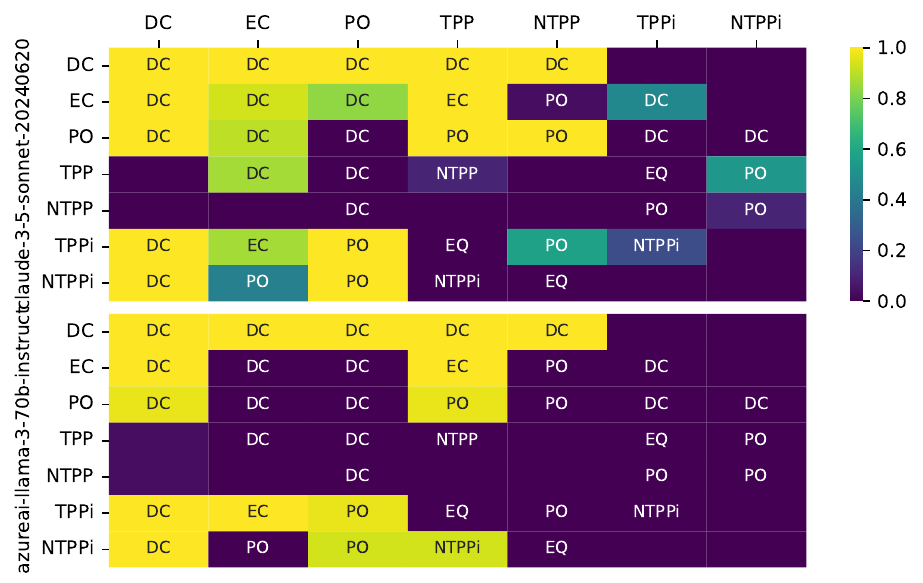} % Reduce the figure size so that it is slightly narrower than the column. Don't use precise values for figure width.This setup will avoid overfull boxes.
\caption{RCC8 preferred relations shaded by mean Jaccard Index (n=30 repeats) for the best performing model, \claude\ and the worst performing model \llama. Labels show preferred relations as reported by Ragni et al \shortcite{ragni2007cross}.
 }
\label{ct-jaccard-pref}
\end{figure}

Fig. \ref{ct-jaccard-pref} shows the \jaccard\ for each cell for  the best performing model, \claude\ and the worst performing model \llama, with actual predictions made by each model in the appendix, Table \ref{rcc8preferred}.

Some responses contained a clear preference with a good argument (E.g. Appendix Example \ref{answer-2} where \claude's response of  \DC\ is indeed the preference reported by Ragni et al \shortcite{ragni2007cross}. 
% 
%
% \textbf{system:}
% \textit{You are a helpful assistant. Consider the following set of eight pairwise disjoint and mutually exhaustive binary spatial relations, between two one piece spatial 2D regions. These relations form part of the well known RCC-8 qualitative spatial reasoning calculus. DC(x,y) means that x and y are disconnected and share no spatial parts. EC(x,y) means that x and y touch at a boundary but do not share any interior parts. PO(x,y) means that x and y share a spatial part, but neither is part of the other.  TPP(x,y) means that x is part of y and touches y’s boundary. NTPP(x,y) means that x is part of y but does not touch y’s boundary.  TPPi(x,y) is the same as TPP(y,x).  NTPPi(x,y) is the same as NTPP(y,x).  I will now ask you a question about these relations. It is possible that in some cases there could be more than one relation that might hold;  in these cases just give me your single most preferred relationship. Answer the question and provide the final answer in the form: "\#\#\# Answer:"}
%
A more problematic example, also from \claude\  (Appendix Example \ref{answer-3}), yields a confused explanation and an incorrect answer of \EC, when only \PO, \TPP\ or \NTPP\ are possible.

% In its responses, sometimes \chat\  gave a clear preference for one relation, for example: 
% \begin{quote}
%    {\bf Prompt:} \emph{If DC(x,y) and DC(y,z) then what is  your preferred relationship between x and z?}
% {\bf Response:} \emph{
% The relations DC(x,y) and DC(y,z) mean that region x is disconnected from region y and region y is disconnected from region z. These relationships don't provide any information about the spatial relationship between x and z. x and z could be disconnected, they could touch at a boundary, or they could partially overlap, among other possibilities.
% However, in absence of any other information and considering spatial continuity, I would say the most likely relation is DC(x, z), assuming x and z are also disconnected. Please note that this is just a statistical assumption; it's possible for x and z to have any of the other relations defined in RCC-8.}
% \end{quote}

In some cases, models justify their choice by saying it was ``cautious'',\ \pct{22}{8820}, or ``safest'' \pct{117}{8820}. The humans in the Ragni et al \shortcite{ragni2007cross} experiment were not asked to justify their choices, but the claim made by the authors of that paper is that choices were based on computational/simplicity considerations -- which differs from the reasons claimed here.

% In other cases \chat\  justified its choice by saying it was the ``cautious'' choice, or the ``safest choice''. The humans in the \cite{ragni2007cross} experiment were not asked to justify their choices, but the claim made by the authors of that paper noted above is that choices were based on computational/simplicity considerations -- which differs from the reasons claimed here by \chat.

%%

As can be seen in Appendix Table \ref{rcc8preferred},
\claude\ agreed with the average human preference in \pct{21}{37} of cases and only chose two impossible relations.
\gpto\ agreed with the average human preference in \pct{19}{37}, but chose six impossible relations.
\gptt\ agreed with the average human preference in  \pct{18}{37} but chose 17 impossible relations. Whilst better than guessing ($\approx 13\%$, these results do not show that LLM preferences are aligned with human preferences.

\begin{figure}[ht]
\centering
\includegraphics[width=0.99\columnwidth]{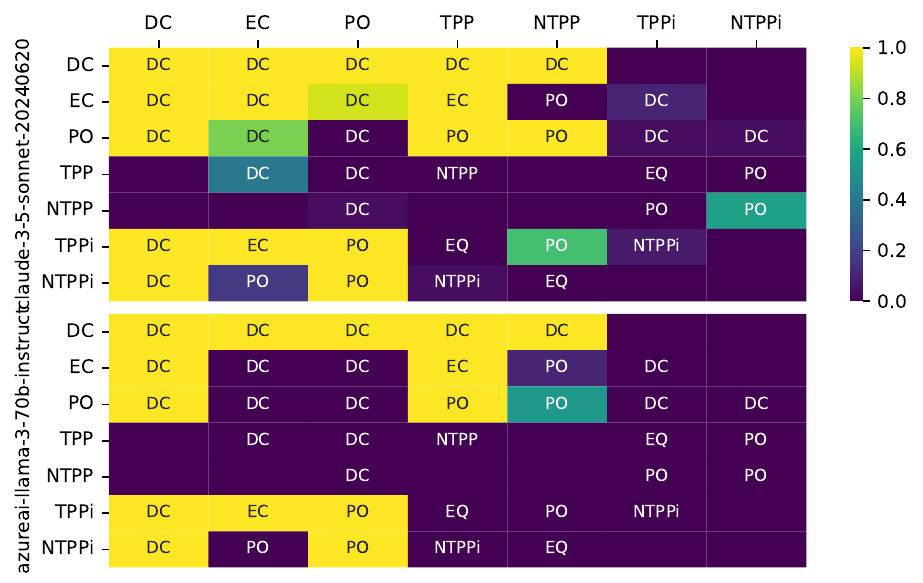} % Reduce the figure size so that it is slightly narrower than the column. Don't use precise values for figure width.This setup will avoid overfull boxes.
\caption{Same as Figure \ref{ct-jaccard-pref} but for anonymous.
 }
\label{xct-jaccard-pref}
\end{figure}

In the anonymous version of this experiment \gpto\ and \llama\ had similar performance with both the eponymous and anonymous relations, but all other models had worse performance with anonymous relations (Table \ref{overall-summary}).

% As can be seen in Table \ref{xrcc8preferred},
% \claude\ agreed with the average human preference in \pct{20}{37},
% \gpto\ in \pct{18}{37},
% and \gptt\ in \pct{16}{37} cases.

% \chat\ only agreed with the average human on 24/49 (48.99\%) though in a further five cases it agreed with one of the language groups (twice times Mongolian, twice German), and in one case the Mongolian preference was equal between the overall choice and the choice \chat\  made giving a an agreement with the German speakers of 26/49 (53.06\%) and 27/49 (55.1\%) with the Mongolian speakers.  In three cases (6.12\%) it actually predicted an impossible relation as its preferred composition (this also happened sometimes in the human experiment), and in two cases (out of 11) where t    he composition is unique it failed to note that. In all the other cases, 20/49 (60.82\%) its preferred relation was possible but not preferred by either language group or overall.   

% The overall agreement with humans in this anonymised case is actually significantly higher than the non-anonymised experiment above; it is not clear if there is a reason for this and if so what.\rob{Is this still true? -NO}

% ***************** say something about whether this is better or worse than the full composition prediction case).

\section{Experiment 3 : Spatial Continuity}
Continuity networks were introduced by Randell and Cohn \shortcite{randell1989modelling} to represent the set of possible `next’ relations that might apply, assuming that motion is continuous and any transformations in the shape and/or size of an object are also continuous.  Subsequently these have been termed ``conceptual neighbourhoods’’ \cite{freksa1992temporal}.  Fortunately, the abbreviation for both of these terms is CN, which we shall use henceforth.  The CN for RCC-8 is depicted in Fig. \ref{rcc8-cn-diagram}.  
% The system prompt was the same as for the CT prompt above, but with two additional sentences at the end: \emph{ In all these questions, assume that all motion is continuous, and any deformations of the spatial regions are also continuous. Also please describe the motion  or changes of size or shape that would be required for the change of relation to take place}. 
Each prompt emphasises the need for the relation to be an immediate next relation.   
Examples of  \claude\  correct and incorrect prompt/response pairs  can be found in the Appendix (correct: Example \ref{answer-4} for  \DC; incorrect Example \ref{answer-5}, where \claude\ correctly predicted \PO, and \EQ\ as neighbours of \TPPi\, but incorrectly predicted \EC, \DC, \NTPP\  and \TPP\  and missed \NTPPi).

\gptf\ performed best,
%($0.67 \pm 0.029$, 
see Table \ref{overall-summary}  column 3epon for all the LLM results.
It is immediately noticeable that the generated tables are not symmetric, although they should be (since all the arrows in Fig. \ref{rcc8-cn-diagram} are bi-directional).

\begin{table}[htb]
\centering
\noindent\resizebox{\columnwidth}{!}{
\input{table-rcc8-continuity-results}

}
\caption{The Continuity Table for RCC-8 produced by our best performing LLM (\gptf) and the worst performing LLM (\gptt). An `x' means that the  relation in that column is predicted as an immediate neighbour of the relation in that row.  An empty box means that the relation is not predicted as an immediate neighbour.  Green means that the prediction was correct and red that it was incorrect. For each model and question, we take the most commonly occurring answer across thirty repetitions. }
\label{rcc8CN}
\end{table}
%%%%%%%%%%%%%%%%%%%%%%%%%%%%%%%%%%%%%%%%%%%%%%%%%%%%%%%%%%%%%%%%%%%%%%%%%%%%%%%%

% \begin{figure}[htb]
% \centering
% %\includegraphics[width=0.99\columnwidth]{RCC8-CN-table.pdf} % Reduce the figure size so that it is slightly narrower than the column. Don't use precise values for figure width.This setup will avoid overfull boxes.
% \caption{The Continuity Table for RCC8 produced by \chat. An `x' means that the  relation in that column is predicted as an immediate neighbour of the relation in that row.  An empty box means that the relation is not predicted as an immediate neighbour.  Green means that the prediction was correct and red that it was incorrect. The leading diagonal is white since a relation is not a next relation of itself.}
% \label{rcc8CN}
% \end{figure}

% In order to test whether the result was influenced by prior knowledge of RCC-8 gained as part of its training, we also performed the same experiment, but with all the relation names anonymised to disguise the connection to RCC-8. 
% The prompt was the same as above except for the change of relation names and the omission of the second sentence. 

In the anonymous version of this experiment
\claude\ performed best,
%($0.61 \pm 0.026$, 
Table \ref{overall-summary}, column 3anon). The results are given in Table \ref{xrcc8CN}. It is  striking how much worse all models except \claude\ do in the anonymous case compared to the eponymous case, suggesting that they are more of a ``parrot'' \cite{bender2021dangers} than a reasoner, though we suspect that the RCC-8 neighbours are rarely, if ever explicitly stated in textual form in the literature.

\begin{table}[htb]
\centering
\noindent\resizebox{\columnwidth}{!}{
\input{table-xrcc8-continuity-results}

}
\caption{The Continuity Table for RCC-8 produced by two LLMs using anonymised relation names. The meaning of the colouring is the same as in Table \ref{rcc8CN} }
\label{xrcc8CN}
\end{table}
%%%%%%%%%%%%%%%%%%%%%%%%%%%%%%%%%%%%%%%%%%%%%%%%%%%%%%%%%%%%%%%%%%%%%%%%%%%%%%%%

\section{Summary }

%%%%%%%%%%%%%%%%%%%%%%%%%%%%%%%%%%%%%%%%%%%%%%%%%%%%%%%%%%%%%%%%%%%%%%%%%%%%%%%%
\begin{table*}[ht]
\centering
\noindent\resizebox{\textwidth}{!}{
\input{table-summary-results}
}
\caption{Comparison of model performance 
%by experiment
based on the mean \jaccard\  
for each question and answer.  Entries in bold
are the best performing model in each experiment. Uncertainty is expressed as the 95\% confidence interval, n=30 repeats.}
\label{overall-summary}
\end{table*}
%%%%%%%%%%%%%%%%%%%%%%%%%%%%%%%%%%%%%%%%%%%%%%%%%%%%%%%%%%%%%%%%%%%%%%%%%%%%%%%%

The best performing models overall were 
\claude\ %(0.60 $\pm$ 0.016) 
and 
\gptf\ %(0.53 $\pm$ 0.0022) 
(Table \ref{overall-summary}) with \llama\ and \gptt\ noticeably less performant.
%( 0.37 $\pm$ 0.013 and 0.35 $\pm$ 0.021).
%
LLMs are non-deterministic and experiment repeats are essential to quantify the variability of answers.

\begin{figure}[ht]
\centering
\includegraphics[width=0.99\columnwidth]{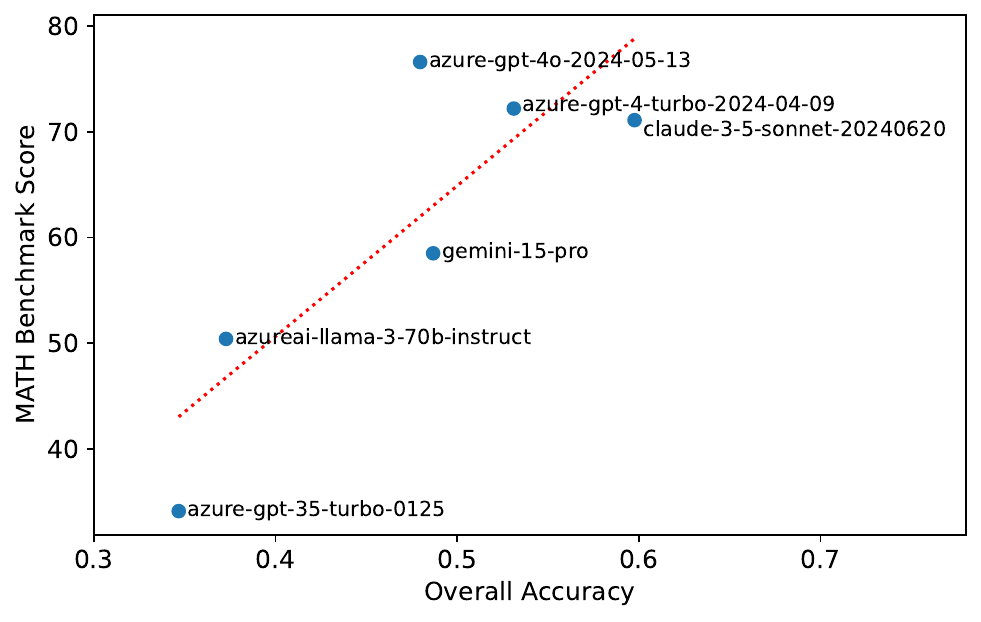} % Reduce the figure size so that it is slightly narrower than the column. Don't use precise values for figure width.This setup will avoid overfull boxes.
\caption{Overall accuracy compared to the MATH benchmark \cite{hendrycks2021measuring}. MATH data courtesy vellum.ai, retrieved 8-Aug-2024.
 }
\label{math}
\end{figure}

Although \claude\ is the newest and best performing model overall and GPT 4 series models outperform \gptt, \gpto\ was released one month after \gptf\ and has worse performance in experiments 2 and 3 suggesting that newer models are not necessarily better than older models in the same family. Although \gptf\ and \gpto\ are both GPT 4 models, it is evident that \gpto\ has had different alignment training and changes to its underlying  engineering\footnote{%See 
\url{https://openai.com/index/hello-gpt-4o/} accessed 10-Aug-24}.
Our results are similar to those from the MATH benchmark \cite{hendrycks2021measuring} (Fig. \ref{math}), except that \gpto\ performs noticeably better in MATH than in our experiments.

% claude-3-opus
% performed best overall with experiments comprising eponymous relations and
% gpt-4-turbo performing best with anonymised relations.  gpt-4-turbo actually also outperformed itself in two of the experiments when dealing with anonymised relations (Experiments 1 and 3).  All other models always performed better on eponymous relations than anonymised relations.  This suggests that in general the models were exploiting general knowledge about RCC-8 from their training data, performing less well when this was irrelevant in the anonymised experiments, whilst gpt-4-turbo seems to have more facility to ``reason from first principles" and not being ``corrupted'' by poorly memorised training data. gpt-35-turbo
% performs  relatively well in experiment 1 with eponymous relations, but poorly in
% other experiments; a possible reason for this is that experiment 1 concerns compositional reasoning which is likely to have appeared much more in the training data than the preferred relations of Experiment 2, or the continuity based reasoning of Experiment 3.  
%is arguably more of a knowledge-based, lookup task than a reasoning task. 
It is noticeable that models seem to rarely miss predicting \DC\  in Experiment 1 and \DC\   is often a human preferred relation in Experiment 2 which is similarly well predicted by the LLMs. This suggests that \DC\  is perhaps the easiest relation for LLMs to reason about.  It can also be seen, from the bottom row of Table \ref{overall-summary}, that Experiment 3 was the easiest reasoning task for the LLMs on average with both the eponymous and anonymous average scores being better than the corresponding score for the other two experiments.

All models except \claude\ and \llama\ perform better with spatial continuity than composition (Table \ref{overall-summary}). 

\vspace*{-.1in}

\section{Concluding Remarks and Future Work}

%This investigation has supported
The results support the widely-held view that LLMs can struggle to do reasoning tasks\footnote{ Bender et al \shortcite{bender2021dangers} have observed that LLMs might be regarded just as ``stochastic parrots'' and thus it is not surprising that  logically correct deductive reasoning is challenging for an LLM.}. In Experiment 1, where LLMs  were asked to compute the entire CT for RCC-8, this is a non-trivial task even for humans, so it is perhaps not surprising that LLMs fell well short of 100\% accuracy.

%-- the scores of 71.94\% (and 67.09\% for the anonymised relations) are clearly much better than chance and do suggest a reasonable facility to perform such computations.   
A detailed analysis of the actual conversations in the supplementary material show that sometimes LLMs do appear able to do some interesting (qualitative) spatial reasoning, but often fail, sometimes making elementary mistakes, and showing confusion about what the task was.  They also show inconsistency in being able to reason correctly about a relation but not its inverse.  
%They also sometimes confuse a relation with its inverse. 
It is possible that fine tuning, explicit chain-of-thought prompting, or more carefully engineered prompts might improve performance; however, given the stochastic nature of LLMs it seems unlikely  that the results would be as good as logical reasoning (Experiment 2 on preferred relations is of course not strictly a logical reasoning exercise, except for the requirement not to predict spatially impossible relations).

There are a variety of avenues for further work which present themselves. Other calculi could be investigated  -- for example the coarser calculus RCC-5\footnote{Preliminary results for the eponymous RCC-5 CT  in the appendix  suggest that LLMs find RCC-5 easier than RCC-8.}, or calculi for reasoning about direction or size \cite{cohn2008qualitative}. Other LLMs could be evaluated -- though since  new LLMs and new LLM versions are continually being released, this is a challenge with no definite stopping point. Tracking the change in performance of a particular LLM across releases would also be of interest -- though in the case of closed LLMs such as the GPT series  where the owners have the right to harvest user conversations and use them for future training, it will not be clear if any improvement is the result of leakage from the previous conversation or more general performance improvement\footnote{However, note that no feedback was given to any of the LLMs investigated as to whether the 
%proffered 
response was correct or not.}.  
%It has already been observed \cite{cohn2023dialectical} that different LLMs have different strengths -- d
Determining which LLMs are better at which spatial reasoning tasks would also be worth of future investigation. The overall conclusion that LLMs in general struggle with more complex spatial reasoning tasks is likely to remain the case, at least for the foreseeable future. As already noted above, even with temperature set to zero, LLMs are not deterministic, different temperatures in the LLM APIs could be tried to see if settings other than the default are reliably better. 
%Different prompts and prompting strategies could be tried, though arguably if QSR has is viewed as a form of commonsense reasoning, it should not be necessary to devise specific prompts to elicit commonsense behaviour. 

 Experiment 2 above  already investigated how LLM performance compared to human performance to a limited extent but further investigation would be worthwhile, including a head-to-head comparison rather than simply taking a result from the literature originally intended to investigate a different question.  Another interesting avenue for further work will be to explore the use of multimodal  models -- when humans perform spatial reasoning tasks including the challenge of building a CT, it is natural to use pencil and paper to sketch diagrams and possible scenarios -- investigating whether a multi-modal model with such abilities (including the ability to analyse its own drawings) would be of great interest to the spatial reasoning community.
%
 %As already noted above, the 
 The LLMs failed to produce symmetric CTs, partly as a result of apparent bias against the inverse relations.  The extent and reasons for this 
 %table asymmetry
 warrants further investigation and analysis. 

 As mentioned above, another possible avenue of research is to investigate different prompting strategies, including k-shot \cite{dang2022prompt}, chain-of-thought \cite{wei2022chain} and tree-of-thought \cite{yao2023tree}. Not doing so was deliberate in this paper as  we wanted to establish a set of base-line results before considering other prompting strategies.  
 %Also, if QSR has is viewed as a form of commonsense reasoning, it should not be necessary to devise specific prompts to elicit commonsense behaviour. 
%
In this paper we use the mean \jaccard\ across 30 repeats of each answer to measure answer accuracy and take the mean of the means across a set of answers to measure model accuracy. There are other methods which could be investigated, though it is already clear the LLMs are far from good reasoners for RCC-8. one alternative is to compute the modal \jaccard\  for each RCC-8 relation individually and average those for each cell or one could use an alternative error metric such as the  \citet{Levenshtein1965BinaryCC} or Edit distance. One could also take account of the distance between relations in the CN and treat deviations which are further from the correct response more harshly (e.g. if the single correct response was \DC\  and a model predicted \EC\ then that would have a smaller penalty than if it predicted \PO\  -- cf \cite{bruns1996similarity}).

Although the results presented here are comparable to the MATH benchmark, there are some notable differences with the GPT 4 series of models, and we suggest that a separate comprehensive QSR benchmark be established.
 
 % It would also be useful to be able to automate the scoring -- we tried different prompts to force the LLMs to just answer the question (i.e. predict the relation(s) requested) but it proved impossible to enforce this, so manual inspection was required to ascertain the correctness of the answer provided.
 %I was interested in exploring in how the ``vanilla'' LLM would perform. Whilst for specific downstream tasks, fine-tuning or employing specific prompting strategies may reasonable, there is an argument to be made that for commonsense reasoning, this is not a reasonable strategy since the task is a general one rather than a specific downstream task.

% it is worth noting that although the author did not use the LLM's feedback mechanism to indicate if the response was correct or not, there is still the risk that the LLM developers will use the interactions to train the model so that the experiments might not be replicable.
%\appendix

% \section{Data statement}
% All the interactions with the LLMs  that support the summary tables can be found in the supplementary material which will be made available online if accepted. 

\section{Data Access}

The data associated with this paper are available from the following
GitHub repository: \url{https://github.com/RobBlackwell/can-llms-reason-about-the-rcc}
Note that the data are encrypted to avoid the questions and answers
being automatically harvested to become LLM training data. However, the decryption procedure and
password are included in the README file.
 
\section{Funding}

This work was supported by the Fundamental Research priority area of
The Alan Turing Institute.

AGC thanks the Turing’s Defence and Security programme
through a partnership with the UK government in accordance with the
framework agreement between GCHQ and The Alan Turing Institute, and
for support provided by the Economic and Social Research Council
(ESRC) under grant ES/W003473/1.

We thank Microsoft Research - Accelerating Foundation Models Research
program, for the provision of Azure resources to access OpenAI models.

\section{Contribution Statement}
AC conceived the original idea. RB implemented the benchmark in consultation with AC and performed the evaluations. AC wrote the first draft and both authors contributed to subsequent drafts.

\bibliography{qr}

\section{Appendix}

This appendix provides supplementary information and gives examples of interactions  which  could not fit into the main paper -- instead the the main paper refers to this appendix.  The appendix also contains additional figures and tables which would not fit into the main paper, giving and comparing the performance of the different LLMs.  
% Finally there is a section giving fuller answers to the reproducibility checklist.
Benchmark questions and results are available in a GitHub repository at https://github.com/RobBlackwell/can-llms-reason-about-the-rcc.
\subsection{Example answers referred to in the paper}
\begin{Example}
{\bf \claude's first answer to \DC{\tiny$^\circ$}\DC }
\\

 \label{answer-1}
\begin{quote}
\noindent\textbf{assistant:} 
{\itshape To answer this question, let's consider the given information and the possible relationships between x and z:
\lf 
\lf 
\lf 
1. We know that DC(x,y) - x and y are disconnected
\lf 
2. We also know that DC(y,z) - y and z are disconnected
\lf 
\lf 
\lf 
Given these conditions, x and z could have any of the following relationships:
\lf 
\lf 
\lf 
1. DC(x,z) - x and z could be disconnected
\lf 
2. EC(x,z) - x and z could be externally connected
\lf 
3. PO(x,z) - x and z could partially overlap
\lf 
4. TPP(x,z) - x could be a tangential proper part of z
\lf 
5. NTPP(x,z) - x could be a non-tangential proper part of z
\lf 
6. TPPi(x,z) - z could be a tangential proper part of x
\lf 
7. NTPPi(x,z) - z could be a non-tangential proper part of x
\lf 
8. EQ(x,z) - x and z could be equal
\lf 
\lf 
\lf 
The reason all relationships are possible is that the disconnection of both x and y from z does not constrain the relationship between x and z in any way. They could be completely separate or in any other spatial configuration.
\lf 
\lf 
\lf 
\#\#\# Answer:
\lf 
DC(x,z), EC(x,z), PO(x,z), TPP(x,z), NTPP(x,z), TPPi(x,z), NTPPi(x,z), EQ(x,z)
}
\end{quote}
\end{Example}

\begin{Example}
{\bf  One instance of \claude's preferred answer to \DC{\tiny$^\circ$}\DC }
\label{answer-2}

\systemprompt{...}{It is possible that in some cases there could be
more than one relation that might hold;  in these cases just give me your single most preferred relationship.}{...}

\begin{quote}
\noindent\textbf{user:} \textit{If DC(x,y) and DC(y,z) then what is  your preferred relationship between x and z? }

\noindent\textbf{assistant:} 
\begin{itshape}
To answer this question, we need to consider the implications of the given information and the possible relationships between x and z.
\lf
\lf
\lf
Given:
\lf
- DC(x,y): x and y are disconnected and share no spatial parts
\lf
- DC(y,z): y and z are disconnected and share no spatial parts
\lf
\lf
\lf
When we consider the relationship between x and z, we can deduce that:
\lf
\lf
\lf
1. x and z could be completely separate from each other, just as they are both separate from y.
\lf
2. x and z could be touching at their boundaries.
\lf
3. x and z could be overlapping or even one could be part of the other.
\lf
\lf
\lf
However, since we are asked for the preferred relationship, we should consider the most likely scenario based on the given information.
\lf
\lf
\lf
The fact that both x and z are disconnected from y suggests that they are likely to be separate from each other as well. While it's possible for them to have a different relationship, the most conservative and likely assumption based on the given information is that x and z are also disconnected.
\lf
\lf
\lf
Therefore, the preferred relationship between x and z is DC(x,z).
\lf
\lf
\lf
\#\#\# Answer: DC(x,z)
\end{itshape}
\end{quote}

\end{Example}

\begin{Example}
{\bf An answer from \claude\ to \EC{\tiny$^\circ$}\NTPP }
\label{answer-3}
\begin{quote}

\noindent\textbf{user:} \textit{If EC(x,y) and NTPP(y,z) then what is your preferred relationship between x and z?}

\noindent\textbf{assistant:} 
{\itshape
To answer this question, let's consider the relationships given:
\lf
\lf
\lf
1. EC(x,y): x and y are externally connected, meaning they touch at their boundaries but don't share interior parts.
\lf
2. NTPP(y,z): y is a non-tangential proper part of z, meaning y is completely inside z without touching z's boundary.
\lf
\lf
\lf
Given these relationships, we need to determine the most likely relationship between x and z.
\lf
\lf
\lf
Since y is completely inside z (NTPP), and x is only touching y at its boundary (EC), x must be outside of z. However, there are two possibilities for how x and z might relate:
\lf
\lf
\lf
1. DC(x,z): x and z could be disconnected, with some space between them.
\lf
2. EC(x,z): x could be touching the boundary of z.
\lf
\lf
\lf
Of these two options, EC(x,z) is more likely because we know that x is touching y, which is inside z. For x to be disconnected from z, there would need to be some space between y and z's boundary, which is not guaranteed by the given information.
\lf
\lf
\lf
Therefore, the preferred relationship between x and z is EC(x,z).
\lf
\lf
\lf
\#\#\# Answer: EC(x,z)
}

\end{quote}

\end{Example}

\begin{Example}
\label{answer-4}
{\bf An example correct response from \claude\ when asked about the conceptual neighbours of \DC. }
\begin{quote}
\systemprompt{...}{In the question, assume that all motion is continuous, and any deformations of the spatial regions are also continuous. Also please describe the motion or changes of size or shape that would be required for the change of relation to take place.}{...}
% {\bf system:}
% {\itshape
% You are a helpful assistant. Consider the following set of eight pairwise disjoint and mutually exhaustive binary spatial relations. These relations form part of the well known RCC-8 qualitative spatial reasoning calculus. DC(x,y) means that x and y are disconnected and share no spatial parts. EC(x,y) means that x and y touch at a boundary but do not share any interior parts. PO(x,y) means that x and y share a spatial part, but neither is part of the other.  TPP(x,y) means that x is part of y and touches y’s boundary. NTPP(x,y) means that x is part of y but does not touch y’s boundary.  TPPi(x,y) is the same as TPP(y,x).  NTPPi(x,y) is the same as NTPP(y,x).  Finally, EQ(x,y) means that x and y are coincident. I will now ask you a question about these relations. In the question, assume that all motion is continuous, and any deformations of the spatial regions are also continuous. Also please describe the motion or changes of size or shape that would be required for the change of relation to take place. Answer the question and provide the final answer in the form: \texttt{"}\#\#\# Answer:\texttt{"}
% }

\noindent{\bf user:}\textit{If DC(a,b) holds, then what are the next possible relations that might hold as the two regions translate or change their shape or size?  I am interested in the immediate next relation, so no intermediate relation must hold.}

\noindent{\bf assistant:}
{\itshape
To answer this question, let's consider the possible changes that could occur to regions a and b starting from the DC (disconnected) relation:
\lf
\lf
\lf
1. Translation: If either region a or b (or both) move towards each other, the next possible relation would be EC (externally connected). This occurs when the regions just touch at their boundaries.
\lf
\lf
\lf
2. Change in shape: If either region a or b (or both) change shape in a way that extends towards the other region, they could also reach the EC relation when their boundaries just touch.
\lf
\lf
\lf
3. Change in size: If either region a or b (or both) grow in size towards each other, they could again reach the EC relation when their boundaries just touch.
\lf
\lf
\lf
It's important to note that no other relation can be reached directly from DC without first passing through EC. This is because:
\lf
\lf
\lf
- To reach PO (partially overlapping), the regions would first need to touch (EC) before overlapping.
\lf
- To reach TPP or NTPP (tangential or non-tangential proper part), one region would need to first touch the other (EC) before becoming part of it.
\lf
- TPPi and NTPPi are inverse relations of TPP and NTPP, so they also cannot be reached directly from DC.
\lf
- EQ (equal) cannot be reached directly from DC as the regions would need to touch and then perfectly align, which would involve intermediate relations.
\lf
\lf
\lf
Therefore, the only immediate next relation possible from DC is EC.
\lf
\lf
\lf
\#\#\# Answer:
\lf
The only next possible relation that can immediately follow DC(a,b) is EC(a,b). This can occur through translation of one or both regions towards each other, or through growth or shape change of one or both regions until their boundaries just touch.
}
\end{quote}
\end{Example}

\begin{Example}
\label{answer-5}
{\bf An example incorrect response from \claude\ when asked about the conceptual neighbours of \TPPi. }

    \begin{quote}
\noindent{\bf user:}\textit{If TPPi(a,b) holds, then what are the next possible relations that might hold as the two regions translate or change their shape or size?  I am interested in the immediate next relation, so no intermediate relation must hold.}

\noindent{\bf assistant:}
{\itshape
To answer this question, let's consider the initial state where TPPi(a,b) holds, meaning that b is a tangential proper part of a. This means that b is completely contained within a, and touches a's boundary at least at one point. From this state, we can consider the possible immediate next relations based on continuous motion, translation, or changes in shape or size.
\lf
\lf
\lf
The possible next relations are:
\lf
\lf
\lf
1. PO(a,b): If b grows or a shrinks in such a way that part of b extends outside of a while still maintaining some overlap, the relation would change to partial overlap.
\lf
\lf
\lf
2. EC(a,b): If a shrinks precisely along its boundary with b, or if b grows exactly to match a's boundary, they would become externally connected.
\lf
\lf
\lf
3. DC(a,b): If a rapidly shrinks away from b or b rapidly shrinks away from a's boundary, they could become disconnected. This is less likely in a continuous motion scenario but technically possible.
\lf
\lf
\lf
4. EQ(a,b): If a shrinks or b grows so that their boundaries exactly match, they would become equal.
\lf
\lf
\lf
5. NTPP(a,b): If b grows to completely engulf a, including a's boundary, the relation would flip to a being a non-tangential proper part of b.
\lf
\lf
\lf
6. TPP(a,b): If b grows to completely contain a but still touches a's boundary at least at one point, the relation would flip to a being a tangential proper part of b.
\lf
\lf
\lf
It's important to note that the relation cannot directly transition to NTPPi(a,b) from TPPi(a,b) without passing through an intermediate state, as this would require b to first separate from a's boundary before becoming properly contained within a.
\lf
\lf
\lf
\#\#\# Answer: The next possible relations that might immediately hold after TPPi(a,b) are PO(a,b), EC(a,b), DC(a,b), EQ(a,b), NTPP(a,b), or TPP(a,b).
}
\end{quote}

\end{Example}

\subsection{Experiment 1 Supplementary Information}

Table \ref{rcc8-composition-table} is provided in addition to Figure \ref{ct-jaccard} to show specific errors of omission and commission for the most commonly occurring answers.

%%%%%%%%%%%%%%%%%%%%%%%%%%%%%%%%%%%%%%%%%%%%%%%%%%%%%%%%%%%%%%%%%%%%%%%%%%%%%%%%
\begin{table*}[htb]
\centering
\noindent\resizebox{\textwidth}{!}{
\input{table-rcc8-composition-results}

}
\caption{The CTs for RCC-8 produced by our best performing LLM (\claude) and the worst performing LLM (\gptt). The entry in each cell uses the following coding: {\sf D (DC), E (EC), P (PO), T (TPP), N (NTPP), t (TPPi), n (NTPPi), Q (EQ).} For each model and composition, we take the most commonly occurring answer across 30 repetitions. Black means that relation is correctly predicted, red  that relation is incorrectly predicted (also indicated with a $-$), blue  that the relation was incorrectly missing (also indicated with a $+$).}
\label{rcc8-composition-table}
\end{table*}
%%%%%%%%%%%%%%%%%%%%%%%%%%%%%%%%%%%%%%%%%%%%%%%%%%%%%%%%%%%%%%%%%%%%%%%%%%%%%%%%

% Figure \ref{ct-jaccard} is another way of visualising Table \ref{rcc8-composition-table}.

\subsection{Experiment 2 Supplementary Information}

Table \ref{rcc8preferred} is provided in addition to Figure \ref{ct-jaccard-pref} to show preferred relations for the most commonly occurring answers.

%%%%%%%%%%%%%%%%%%%%%%%%%%%%%%%%%%%%%%%%%%%%%%%%%%%%%%%%%%%%%%%%%%%%%%%%%%%%%%
\begin{table}[htb]
\centering
\noindent\resizebox{\columnwidth}{!}{
\input{table-rcc8-preferred-results}

}
\caption{
The table of preferred compositions for RCC-8 produced by our best performing LLM (\claude) and the worst performing  (\llama); each cell indicates 
the most commonly occurring answer across 30 repetitions. Green text means it agreed with the human most likely relation, and blue a possible but not the most human-preferred relation. Red means it chose an impossible relation. 
%Blue means it chose a possible but not human-preferred relation. 
%For each model and composition, we take the most commonly occurring answer across 30 repetitions. 
%Of course, since 
Since the humans in Ragni et al's experiment were not unanimous in their choices, the minority humans also made choices not preferred by the majority. }
\label{rcc8preferred}
\end{table}
%%%%%%%%%%%%%%%%%%%%%%%%%%%%%%%%%%%%%%%%%%%%%%%%%%%%%%%%%%%%%%%%%%%%%%%%%%%%%%%%

% Figure \ref{ct-jaccard-pref} is another way of visualising Table \ref{rcc8preferred}.

\subsection{Variability}

\begin{figure}[htb]
\centering
\includegraphics[width=0.99\columnwidth]{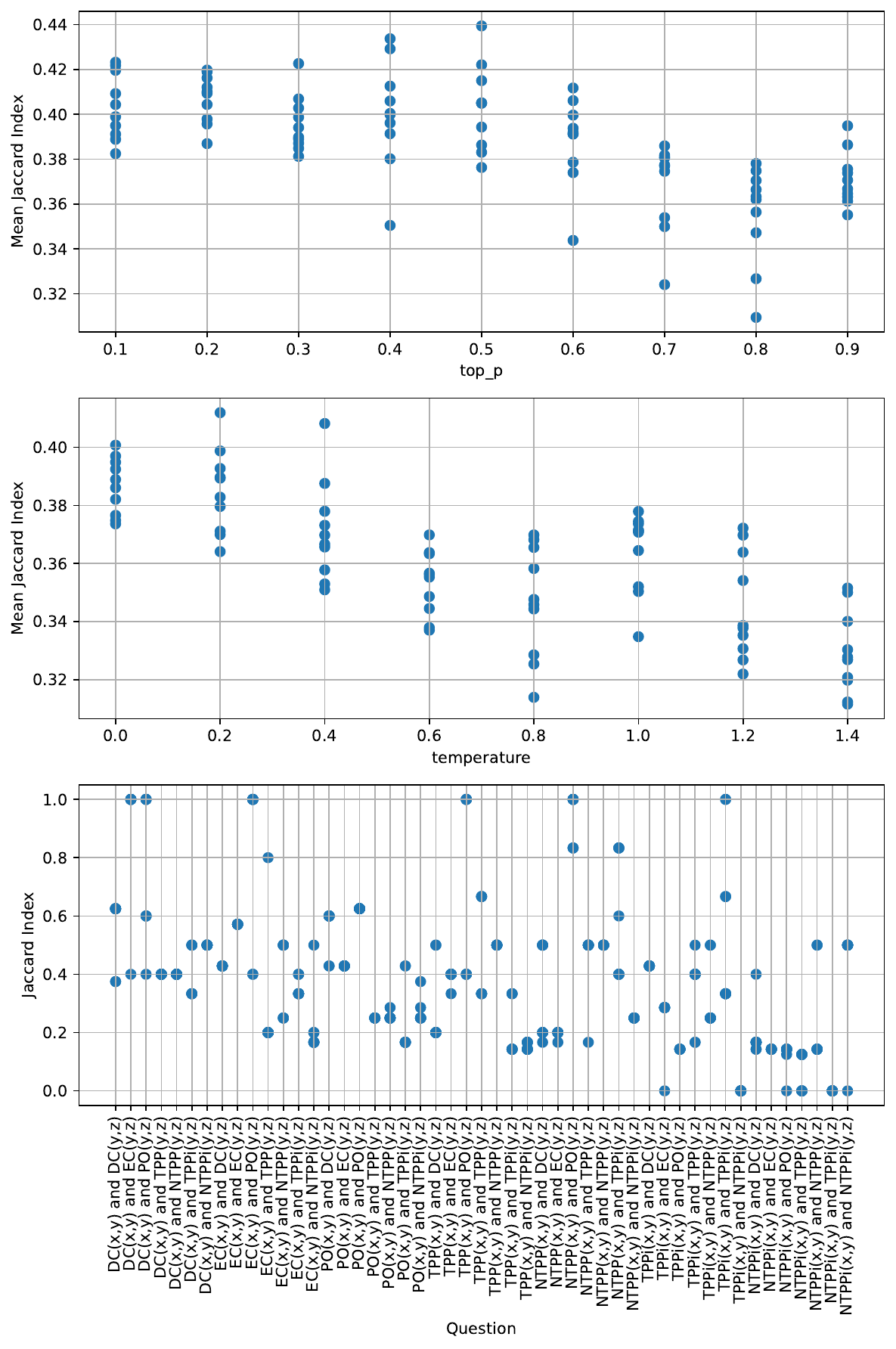} % Reduce the figure size so that it is slightly narrower than the column. Don't use precise values for figure width.This setup will avoid overfull boxes.
\caption{Variability of scores when using \gptt\ i) varying \emph{top\_p}, ii) varying \emph{temperature} and iii) when fixing \emph{temperature} $= 0$. Ten repeats. In all cases we fix the \emph{seed} $=123$.}
\label{rcc8-variability}
\end{figure}

To explore the effect of LLM parameters, we fixed the \emph{seed} and varied \emph{temperature} for \gptt\ applied to our first Experiment 1 question set, and then fixed the \emph{seed} and varied \emph{top\_p}\ (Fig. \ref{rcc8-variability}). Results show considerable variability.

\subsection {Anonymised Experiments}

\begin{figure}[htb]
\centering
\includegraphics[width=0.99\columnwidth]{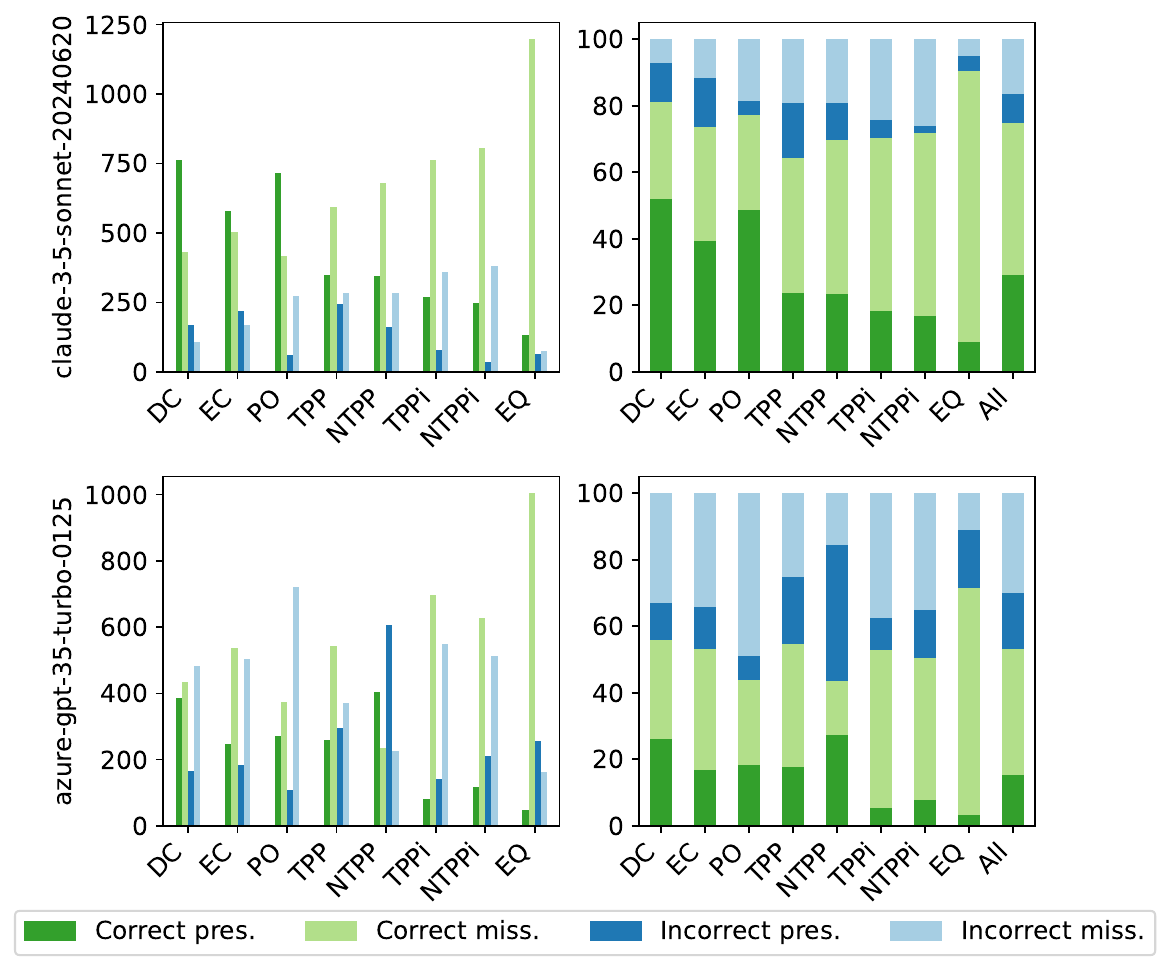} % Reduce the figure size so that it is slightly narrower than the column. Don't use precise values for figure width.This setup will avoid overfull boxes.
\caption{Relation statistics for the CT for RCC-8 with anonymised relations. Layout as for Fig. \ref{ct-relation-statistics}. 
 }
\label{xrcc8-comp-relation-stats}
\end{figure}

The following tables and figures show results of the anonymised experiments for comparison with the eponymous experiments included in the main paper (Table \ref{xrcc8-composition-table}, Figure \ref{xrcc8-comp-relation-stats} and Table \ref{xrcc8preferred}).

%%%%%%%%%%%%%%%%%%%%%%%%%%%%%%%%%%%%%%%%%%%%%%%%%%%%%%%%%%%%%%%%%%%%%%%%%%%%%%%%
\begin{table*}[htb]
\centering
\noindent\resizebox{\textwidth}{!}{
\input{table-xrcc8-composition-results}

}
\caption{The CTs for RCC-8 produced by our best performing LLM (\claude) and the worst performing LLM (\gptt) using anonymised relations. The representation used in the entries is the same as in Table~\ref{rcc8-composition-table}. }
\label{xrcc8-composition-table}
\end{table*}
%%%%%%%%%%%%%%%%%%%%%%%%%%%%%%%%%%%%%%%%%%%%%%%%%%%%%%%%%%%%%%%%%%%%%%%%%%%%%%%%

%%%%%%%%%%%%%%%%%%%%%%%%%%%%%%%%%%%%%%%%%%%%%%%%%%%%%%%%%%%%%%%%%%%%%%%%%%%%%%%%
\begin{table}[htb]
\centering
\noindent\resizebox{\columnwidth}{!}{
\input{table-xrcc8-preferred-results}

}
\caption{The table of preferred compositions for RCC-8 produced by our best performing LLM (\claude) and the worst performing LLM (\llama) LLMs with anonymised relations. The relation names are coloured in the same way as in Table \ref{rcc8preferred}. }
\label{xrcc8preferred}
\end{table}
%%%%%%%%%%%%%%%%%%%%%%%%%%%%%%%%%%%%%%%%%%%%%%%%%%%%%%%%%%%%%%%%%%%%%%%%%%%%%%%%

\subsection{Preliminary Results for RCC-5}

Preliminary results for the RCC-5 CT with eponymous relations are
\claude\ 0.84$\pm$0.011,
\gptf\ 0.72$\pm$0.012,
\gemini\ 0.72$\pm$0.013,
\gpto\ 0.66$\pm$0.014,
\llama\ 0.60$\pm$0.013, and
\gptt\ 0.47$\pm$0.016. All models performed better with RCC-5 than with RCC-8, consistent with RCC-5 being a coarser calculus that is easier to reason about.

\subsection {Computing Infrastructure}

All experiments were conducted on an Apple MacBook Pro with an Apple M2 Pro processor and 32GB of memory. The Large Languages Models themselves
were hosted by suppliers (Google Vertex, Microsoft Azure, Anthropic) and accessed using HTTP APIs as listed in in Table \ref{tab:models}. A \texttt{Makefile} is
provided with each experiment to aid reproducibility. 
% All Python software library versions were recorded in \texttt{requirements.txt}.
All interactions with LLM APIs were recorded in \texttt{answers.jsonl} files including prompts, responses, date and time stamps, API and model versions.

\end{document}

%% file: table-rcc8-continuity-results.tex
\begin{tabular}{|c|c|c|c|c|c|c|c|c|}

\hline
\textbf{azure-gpt-4-turbo-2024-04-09} & \textbf{DC} & \textbf{EC} & \textbf{PO} & \textbf{TPP} & \textbf{NTPP} & \textbf{EQ} & \textbf{TPPi} & \textbf{NTPPi} \\
\hline
\textbf{DC} & \cellcolor{green} & \cellcolor{green}x & \cellcolor{green} & \cellcolor{green} & \cellcolor{green} & \cellcolor{green} & \cellcolor{green} & \cellcolor{green} \\
\hline
\textbf{EC} & \cellcolor{green}x & \cellcolor{green} & \cellcolor{green}x & \cellcolor{green} & \cellcolor{green} & \cellcolor{green} & \cellcolor{green} & \cellcolor{green} \\
\hline
\textbf{PO} & \cellcolor{red}x & \cellcolor{green}x & \cellcolor{green} & \cellcolor{red} & \cellcolor{green} & \cellcolor{green}x & \cellcolor{red} & \cellcolor{green} \\
\hline
\textbf{TPP} & \cellcolor{green} & \cellcolor{red}x & \cellcolor{green}x & \cellcolor{green} & \cellcolor{green}x & \cellcolor{red} & \cellcolor{green} & \cellcolor{green} \\
\hline
\textbf{NTPP} & \cellcolor{green} & \cellcolor{green} & \cellcolor{green} & \cellcolor{green}x & \cellcolor{green} & \cellcolor{green}x & \cellcolor{green} & \cellcolor{green} \\
\hline
\textbf{EQ} & \cellcolor{green} & \cellcolor{green} & \cellcolor{red} & \cellcolor{green}x & \cellcolor{green}x & \cellcolor{green} & \cellcolor{green}x & \cellcolor{green}x \\
\hline
\textbf{TPPi} & \cellcolor{green} & \cellcolor{green} & \cellcolor{red} & \cellcolor{red}x & \cellcolor{green} & \cellcolor{green}x & \cellcolor{green} & \cellcolor{green}x \\
\hline
\textbf{NTPPi} & \cellcolor{green} & \cellcolor{green} & \cellcolor{green} & \cellcolor{green} & \cellcolor{green} & \cellcolor{green}x & \cellcolor{green}x & \cellcolor{green} \\
\hline

\multicolumn{1}{c}{} \\ 

\hline
\textbf{azure-gpt-35-turbo-0125} & \textbf{DC} & \textbf{EC} & \textbf{PO} & \textbf{TPP} & \textbf{NTPP} & \textbf{EQ} & \textbf{TPPi} & \textbf{NTPPi} \\
\hline
\textbf{DC} & \cellcolor{green} & \cellcolor{green}x & \cellcolor{green} & \cellcolor{green} & \cellcolor{green} & \cellcolor{green} & \cellcolor{green} & \cellcolor{green} \\
\hline
\textbf{EC} & \cellcolor{green}x & \cellcolor{green} & \cellcolor{red} & \cellcolor{green} & \cellcolor{green} & \cellcolor{red}x & \cellcolor{green} & \cellcolor{green} \\
\hline
\textbf{PO} & \cellcolor{green} & \cellcolor{red} & \cellcolor{green} & \cellcolor{green}x & \cellcolor{green} & \cellcolor{red} & \cellcolor{green}x & \cellcolor{green} \\
\hline
\textbf{TPP} & \cellcolor{red}x & \cellcolor{red}x & \cellcolor{red} & \cellcolor{green} & \cellcolor{green}x & \cellcolor{green}x & \cellcolor{green} & \cellcolor{green} \\
\hline
\textbf{NTPP} & \cellcolor{green} & \cellcolor{green} & \cellcolor{green} & \cellcolor{green}x & \cellcolor{green} & \cellcolor{green}x & \cellcolor{green} & \cellcolor{green} \\
\hline
\textbf{EQ} & \cellcolor{red}x & \cellcolor{red}x & \cellcolor{red} & \cellcolor{red} & \cellcolor{red} & \cellcolor{green} & \cellcolor{red} & \cellcolor{red} \\
\hline
\textbf{TPPi} & \cellcolor{green} & \cellcolor{green} & \cellcolor{red} & \cellcolor{green} & \cellcolor{red}x & \cellcolor{red} & \cellcolor{green} & \cellcolor{red} \\
\hline
\textbf{NTPPi} & \cellcolor{green} & \cellcolor{green} & \cellcolor{green} & \cellcolor{green} & \cellcolor{green} & \cellcolor{green}x & \cellcolor{green}x & \cellcolor{green} \\
\hline

\end{tabular}

%% file: table-xrcc8-continuity-results.tex
\begin{tabular}{|c|c|c|c|c|c|c|c|c|}

\hline
\textbf{azure-gpt-4-turbo-2024-04-09} & \textbf{DC} & \textbf{EC} & \textbf{PO} & \textbf{TPP} & \textbf{NTPP} & \textbf{EQ} & \textbf{TPPi} & \textbf{NTPPi} \\
\hline
\textbf{DC} & \cellcolor{green} & \cellcolor{green}x & \cellcolor{green} & \cellcolor{green} & \cellcolor{green} & \cellcolor{green} & \cellcolor{green} & \cellcolor{green} \\
\hline
\textbf{EC} & \cellcolor{green}x & \cellcolor{green} & \cellcolor{green}x & \cellcolor{green} & \cellcolor{green} & \cellcolor{green} & \cellcolor{green} & \cellcolor{green} \\
\hline
\textbf{PO} & \cellcolor{red}x & \cellcolor{green}x & \cellcolor{green} & \cellcolor{green}x & \cellcolor{red}x & \cellcolor{green}x & \cellcolor{green}x & \cellcolor{red}x \\
\hline
\textbf{TPP} & \cellcolor{green} & \cellcolor{green} & \cellcolor{green}x & \cellcolor{green} & \cellcolor{green}x & \cellcolor{red} & \cellcolor{green} & \cellcolor{green} \\
\hline
\textbf{NTPP} & \cellcolor{green} & \cellcolor{green} & \cellcolor{green} & \cellcolor{green}x & \cellcolor{green} & \cellcolor{green}x & \cellcolor{green} & \cellcolor{green} \\
\hline
\textbf{EQ} & \cellcolor{green} & \cellcolor{red}x & \cellcolor{green}x & \cellcolor{green}x & \cellcolor{green}x & \cellcolor{green} & \cellcolor{green}x & \cellcolor{green}x \\
\hline
\textbf{TPPi} & \cellcolor{green} & \cellcolor{red}x & \cellcolor{red} & \cellcolor{red}x & \cellcolor{green} & \cellcolor{green}x & \cellcolor{green} & \cellcolor{red} \\
\hline
\textbf{NTPPi} & \cellcolor{green} & \cellcolor{green} & \cellcolor{green} & \cellcolor{green} & \cellcolor{green} & \cellcolor{red} & \cellcolor{green}x & \cellcolor{green} \\
\hline

\multicolumn{1}{c}{} \\ 

\hline
\textbf{azure-gpt-35-turbo-0125} & \textbf{DC} & \textbf{EC} & \textbf{PO} & \textbf{TPP} & \textbf{NTPP} & \textbf{EQ} & \textbf{TPPi} & \textbf{NTPPi} \\
\hline
\textbf{DC} & \cellcolor{green} & \cellcolor{green}x & \cellcolor{red}x & \cellcolor{green} & \cellcolor{green} & \cellcolor{green} & \cellcolor{green} & \cellcolor{green} \\
\hline
\textbf{EC} & \cellcolor{green}x & \cellcolor{green} & \cellcolor{green}x & \cellcolor{green} & \cellcolor{green} & \cellcolor{green} & \cellcolor{green} & \cellcolor{green} \\
\hline
\textbf{PO} & \cellcolor{green} & \cellcolor{green}x & \cellcolor{green} & \cellcolor{red} & \cellcolor{green} & \cellcolor{red} & \cellcolor{red} & \cellcolor{green} \\
\hline
\textbf{TPP} & \cellcolor{red}x & \cellcolor{red}x & \cellcolor{red} & \cellcolor{green} & \cellcolor{red} & \cellcolor{red} & \cellcolor{green} & \cellcolor{green} \\
\hline
\textbf{NTPP} & \cellcolor{green} & \cellcolor{green} & \cellcolor{green} & \cellcolor{red} & \cellcolor{green} & \cellcolor{red} & \cellcolor{green} & \cellcolor{red}x \\
\hline
\textbf{EQ} & \cellcolor{red}x & \cellcolor{red}x & \cellcolor{red} & \cellcolor{red} & \cellcolor{red} & \cellcolor{green} & \cellcolor{red} & \cellcolor{red} \\
\hline
\textbf{TPPi} & \cellcolor{green} & \cellcolor{green} & \cellcolor{red} & \cellcolor{green} & \cellcolor{green} & \cellcolor{red} & \cellcolor{green} & \cellcolor{green}x \\
\hline
\textbf{NTPPi} & \cellcolor{red}x & \cellcolor{red}x & \cellcolor{green} & \cellcolor{green} & \cellcolor{green} & \cellcolor{red} & \cellcolor{red} & \cellcolor{green} \\
\hline

\end{tabular}

%% file: table-summary-results.tex
\begin{tabular}{llllllll}
\toprule
Model & 1epon & 1anon & 2epon & 2anon & 3epon & 3anon & Overall \\
\midrule
claude-3-5-sonnet-20240620 & \textbf{0.69 $\pm$ 0.008} & \textbf{0.58 $\pm$ 0.012} & \textbf{0.56 $\pm$ 0.011} & \textbf{0.54 $\pm$ 0.008} & 0.61 $\pm$ 0.023 & \textbf{0.61 $\pm$ 0.026} & \textbf{0.60 $\pm$ 0.016} \\
azure-gpt-4-turbo-2024-04-09 & 0.50 $\pm$ 0.014 & 0.45 $\pm$ 0.013 & 0.50 $\pm$ 0.014 & 0.49 $\pm$ 0.014 & \textbf{0.67 $\pm$ 0.029} & 0.57 $\pm$ 0.036 & 0.53 $\pm$ 0.022 \\
gemini-15-pro & 0.51 $\pm$ 0.013 & 0.49 $\pm$ 0.011 & 0.51 $\pm$ 0.010 & 0.46 $\pm$ 0.007 & 0.52 $\pm$ 0.021 & 0.42 $\pm$ 0.021 & 0.49 $\pm$ 0.015 \\
azure-gpt-4o-2024-05-13 & 0.47 $\pm$ 0.013 & 0.43 $\pm$ 0.012 & 0.50 $\pm$ 0.012 & 0.50 $\pm$ 0.014 & 0.57 $\pm$ 0.039 & 0.41 $\pm$ 0.033 & 0.48 $\pm$ 0.024 \\
azureai-llama-3-70b-instruct & 0.43 $\pm$ 0.006 & 0.40 $\pm$ 0.010 & 0.40 $\pm$ 0.004 & 0.40 $\pm$ 0.005 & 0.43 $\pm$ 0.018 & 0.18 $\pm$ 0.021 & 0.37 $\pm$ 0.013 \\
azure-gpt-35-turbo-0125 & 0.33 $\pm$ 0.009 & 0.25 $\pm$ 0.011 & 0.44 $\pm$ 0.015 & 0.36 $\pm$ 0.011 & 0.39 $\pm$ 0.034 & 0.30 $\pm$ 0.030 & 0.35 $\pm$ 0.021 \\
\hline
Overall & 0.49 $\pm$ 0.011 & 0.43 $\pm$ 0.012 & 0.49 $\pm$ 0.012 & 0.46 $\pm$ 0.011 & 0.53 $\pm$ 0.028 & 0.42 $\pm$ 0.028 & 0.47 $\pm$ 0.057 \\
\hline
Guess rate & 0.31 $\pm$ 0.012 & 0.31 $\pm$ 0.012 & 0.13 $\pm$ 0.017 & 0.13 $\pm$ 0.017 & 0.26 $\pm$ 0.025 & 0.26 $\pm$ 0.025 & 0.23 $\pm$ 0.033 \\
\bottomrule
\end{tabular}

%% file: table-rcc8-composition-results.tex
\begin{tabular}{|c|c|c|c|c|c|c|c|}

\multicolumn{1}{c}{} \\ 
\hline
\textbf{claude-3-5-sonnet-20240620} & \textbf{DC} & \textbf{EC} & \textbf{PO} & \textbf{TPP} & \textbf{NTPP} & \textbf{TPPi} & \textbf{NTPPi} \\
\hline
\textbf{DC} & D,E,P,T,N,t,n,Q & D,E,P,\textcolor{blue}{+T},\textcolor{blue}{+N} & D,E,P,\textcolor{blue}{+T},\textcolor{blue}{+N} & D,E,P,\textcolor{blue}{+T},\textcolor{blue}{+N} & D,E,P,\textcolor{blue}{+T},\textcolor{blue}{+N} & D,\textcolor{red}{-E} & D \\
\hline
\textbf{EC} & D,E,P,t,n & D,E,P,\textcolor{blue}{+T},\textcolor{blue}{+t},\textcolor{blue}{+Q} & D,E,P,\textcolor{blue}{+T},\textcolor{blue}{+N} & \textcolor{red}{-D},E,P,T,\textcolor{blue}{+N} & \textcolor{red}{-D},\textcolor{red}{-E},P,T,\textcolor{blue}{+N} & D,E,\textcolor{red}{-P} & D \\
\hline
\textbf{PO} & D,E,P,t,n & D,E,P,\textcolor{blue}{+t},\textcolor{blue}{+n} & D,E,P,T,N,t,n,Q & \textcolor{red}{-D},\textcolor{red}{-E},P,T,N & P,T,N & D,E,P,\textcolor{red}{-T},\textcolor{red}{-N},\textcolor{blue}{+t},\textcolor{blue}{+n} & D,E,P,t,n \\
\hline
\textbf{TPP} & D & D,E,\textcolor{red}{-P},\textcolor{red}{-T} & D,E,P,T,N & T,N & \textcolor{red}{-T},N & D,E,P,T,\textcolor{red}{-N},\textcolor{blue}{+t},Q & \textcolor{blue}{+D},\textcolor{blue}{+E},P,\textcolor{red}{-T},\textcolor{red}{-N},\textcolor{blue}{+t},\textcolor{blue}{+n} \\
\hline
\textbf{NTPP} & D & D & D,E,P,T,N & N & N & \textcolor{blue}{+D},\textcolor{blue}{+E},\textcolor{blue}{+P},T,N & \textcolor{blue}{+D},\textcolor{blue}{+E},\textcolor{blue}{+P},\textcolor{blue}{+T},N,\textcolor{blue}{+t},\textcolor{blue}{+n},\textcolor{blue}{+Q} \\
\hline
\textbf{TPPi} & D,E,P,\textcolor{blue}{+t},\textcolor{blue}{+n} & \textcolor{red}{-D},E,P,t,n & \textcolor{red}{-D},\textcolor{red}{-E},P,t,n & P,T,\textcolor{red}{-N},\textcolor{blue}{+t},Q & P,T,N,\textcolor{red}{-Q} & t,n & \textcolor{red}{-t},n \\
\hline
\textbf{NTPPi} & D,E,P,\textcolor{blue}{+t},\textcolor{blue}{+n} & \textcolor{red}{-E},P,\textcolor{blue}{+t},\textcolor{blue}{+n} & P,t,n,\textcolor{red}{-Q} & P,t,n,\textcolor{red}{-Q} & P,T,N,t,n,Q & \textcolor{red}{-t},n & n \\
\hline

\multicolumn{1}{c}{} \\ 
\hline
\textbf{azure-gpt-35-turbo-0125} & \textbf{DC} & \textbf{EC} & \textbf{PO} & \textbf{TPP} & \textbf{NTPP} & \textbf{TPPi} & \textbf{NTPPi} \\
\hline
\textbf{DC} & D,\textcolor{blue}{+E},\textcolor{blue}{+P},\textcolor{blue}{+T},\textcolor{blue}{+N},\textcolor{blue}{+t},\textcolor{blue}{+n},\textcolor{blue}{+Q} & D,E,\textcolor{blue}{+P},\textcolor{blue}{+T},\textcolor{blue}{+N} & D,E,P,\textcolor{blue}{+T},\textcolor{blue}{+N} & D,\textcolor{blue}{+E},\textcolor{blue}{+P},\textcolor{blue}{+T},N & D,\textcolor{blue}{+E},\textcolor{blue}{+P},\textcolor{blue}{+T},N & D,\textcolor{red}{-N} & D,\textcolor{red}{-N} \\
\hline
\textbf{EC} & D,E,\textcolor{blue}{+P},\textcolor{blue}{+t},\textcolor{blue}{+n} & D,E,\textcolor{blue}{+P},\textcolor{blue}{+T},\textcolor{blue}{+t},\textcolor{blue}{+Q} & D,E,P,T,N & \textcolor{red}{-D},\textcolor{blue}{+E},\textcolor{blue}{+P},\textcolor{blue}{+T},N & \textcolor{red}{-D},\textcolor{blue}{+P},\textcolor{blue}{+T},N & D,\textcolor{blue}{+E},\textcolor{red}{-N} & D,\textcolor{red}{-N} \\
\hline
\textbf{PO} & D,E,P,\textcolor{blue}{+t},\textcolor{blue}{+n} & D,E,\textcolor{blue}{+P},\textcolor{blue}{+t},\textcolor{blue}{+n} & D,E,P,T,N,\textcolor{blue}{+t},\textcolor{blue}{+n},\textcolor{blue}{+Q} & \textcolor{blue}{+P},T,N & \textcolor{red}{-D},\textcolor{blue}{+P},\textcolor{blue}{+T},N & D,\textcolor{blue}{+E},\textcolor{blue}{+P},\textcolor{red}{-N},\textcolor{blue}{+t},\textcolor{blue}{+n} & D,\textcolor{blue}{+E},\textcolor{blue}{+P},\textcolor{red}{-N},\textcolor{blue}{+t},\textcolor{blue}{+n} \\
\hline
\textbf{TPP} & D,\textcolor{red}{-N} & D,\textcolor{blue}{+E},\textcolor{red}{-N} & D,\textcolor{blue}{+E},\textcolor{blue}{+P},\textcolor{blue}{+T},N & T,\textcolor{blue}{+N},\textcolor{red}{-Q} & \textcolor{red}{-D},N & \textcolor{blue}{+D},\textcolor{blue}{+E},\textcolor{blue}{+P},T,\textcolor{blue}{+t},Q & D,\textcolor{blue}{+E},\textcolor{blue}{+P},\textcolor{red}{-N},\textcolor{blue}{+t},\textcolor{blue}{+n} \\
\hline
\textbf{NTPP} & D,\textcolor{red}{-N} & D,\textcolor{red}{-N} & D,E,\textcolor{blue}{+P},\textcolor{blue}{+T},\textcolor{blue}{+N} & \textcolor{red}{-D},N & \textcolor{red}{-D},N & \textcolor{blue}{+D},\textcolor{blue}{+E},\textcolor{blue}{+P},T,N,\textcolor{red}{-Q} & \textcolor{blue}{+D},\textcolor{blue}{+E},\textcolor{blue}{+P},\textcolor{blue}{+T},N,\textcolor{blue}{+t},n,\textcolor{blue}{+Q} \\
\hline
\textbf{TPPi} & D,\textcolor{blue}{+E},\textcolor{blue}{+P},\textcolor{red}{-N},\textcolor{blue}{+t},\textcolor{blue}{+n} & \textcolor{red}{-D},\textcolor{blue}{+E},\textcolor{blue}{+P},\textcolor{red}{-N},\textcolor{blue}{+t},\textcolor{blue}{+n} & \textcolor{red}{-D},\textcolor{red}{-E},\textcolor{blue}{+P},\textcolor{blue}{+t},\textcolor{blue}{+n} & \textcolor{red}{-D},\textcolor{blue}{+P},\textcolor{blue}{+T},\textcolor{red}{-N},\textcolor{blue}{+t},\textcolor{blue}{+Q} & \textcolor{red}{-D},\textcolor{blue}{+P},\textcolor{blue}{+T},N & t,\textcolor{blue}{+n},\textcolor{red}{-Q} & \textcolor{red}{-D},\textcolor{red}{-N},\textcolor{blue}{+n} \\
\hline
\textbf{NTPPi} & D,\textcolor{blue}{+E},\textcolor{blue}{+P},\textcolor{red}{-N},\textcolor{blue}{+t},\textcolor{blue}{+n} & \textcolor{red}{-D},\textcolor{red}{-E},\textcolor{blue}{+P},\textcolor{blue}{+t},\textcolor{blue}{+n} & \textcolor{red}{-D},\textcolor{blue}{+P},\textcolor{red}{-N},\textcolor{blue}{+t},\textcolor{blue}{+n} & \textcolor{red}{-D},\textcolor{blue}{+P},\textcolor{red}{-N},\textcolor{blue}{+t},\textcolor{blue}{+n} & \textcolor{red}{-D},\textcolor{blue}{+P},\textcolor{blue}{+T},N,\textcolor{blue}{+t},\textcolor{blue}{+n},\textcolor{blue}{+Q} & \textcolor{red}{-T},\textcolor{red}{-N},\textcolor{blue}{+n} & \textcolor{red}{-D},n \\
\hline

\end{tabular}

%% file: table-rcc8-preferred-results.tex
\begin{tabular}{|c|c|c|c|c|c|c|c|}

\hline
\textbf{claude-3-5-sonnet-20240620} & \textbf{DC} & \textbf{EC} & \textbf{PO} & \textbf{TPP} & \textbf{NTPP} & \textbf{TPPi} & \textbf{NTPPi} \\
\hline
\textbf{DC} & \textcolor{green}{DC} & \textcolor{green}{DC} & \textcolor{green}{EC} & \textcolor{red}{DC} & \textcolor{blue}{EC} & \textcolor{blue}{DC} & \textcolor{green}{DC} \\
\hline
\textbf{EC} & \textcolor{green}{DC} & \textcolor{blue}{PO} & \textcolor{green}{PO} & \textcolor{green}{PO} & \textcolor{green}{DC} & \textcolor{blue}{PO} & \textcolor{blue}{PO} \\
\hline
\textbf{PO} & \textcolor{blue}{DC} & \textcolor{green}{DC} & \textcolor{blue}{PO} & \textcolor{blue}{TPP} & \textcolor{blue}{NTPP} & \textcolor{blue}{TPP} & \textcolor{green}{PO} \\
\hline
\textbf{TPP} & \textcolor{blue}{DC} & \textcolor{green}{DC} & \textcolor{blue}{DC} & \textcolor{blue}{NTPP} & \textcolor{blue}{NTPP} & \textcolor{blue}{NTPP} & \textcolor{blue}{NTPP} \\
\hline
\textbf{NTPP} & \textcolor{blue}{NTPP} & \textcolor{green}{DC} & \textcolor{green}{EC} & \textcolor{green}{PO} & \textcolor{blue}{PO} & \textcolor{green}{DC} & \textcolor{green}{PO} \\
\hline
\textbf{TPPi} & \textcolor{blue}{TPPi} & \textcolor{blue}{NTPPi} & \textcolor{green}{DC} & \textcolor{red}{EC} & \textcolor{green}{PO} & \textcolor{blue}{PO} & \textcolor{blue}{PO} \\
\hline
\textbf{NTPPi} & \textcolor{blue}{NTPPi} & \textcolor{blue}{NTPPi} & \textcolor{green}{DC} & \textcolor{blue}{DC} & \textcolor{blue}{DC} & \textcolor{green}{DC} & \textcolor{green}{DC} \\
\hline

\multicolumn{1}{c}{} \\ 

\hline
\textbf{azureai-llama-3-70b-instruct} & \textbf{DC} & \textbf{EC} & \textbf{PO} & \textbf{TPP} & \textbf{NTPP} & \textbf{TPPi} & \textbf{NTPPi} \\
\hline
\textbf{DC} & \textcolor{green}{DC} & \textcolor{blue}{EC} & \textcolor{green}{EC} & \textcolor{red}{EC} & \textcolor{blue}{EC} & \textcolor{red}{EC} & \textcolor{green}{DC} \\
\hline
\textbf{EC} & \textcolor{blue}{EC} & \textcolor{blue}{PO} & \textcolor{green}{PO} & \textcolor{blue}{NTPP} & \textcolor{green}{DC} & \textcolor{blue}{PO} & \textcolor{red}{NTPP} \\
\hline
\textbf{PO} & \textcolor{blue}{DC} & \textcolor{blue}{EC} & \textcolor{blue}{TPP} & \textcolor{blue}{TPP} & \textcolor{blue}{NTPP} & \textcolor{blue}{TPP} & \textcolor{red}{NTPP} \\
\hline
\textbf{TPP} & \textcolor{blue}{DC} & \textcolor{green}{DC} & \textcolor{red}{NTPP} & \textcolor{blue}{NTPP} & \textcolor{blue}{NTPP} & \textcolor{blue}{NTPP} & \textcolor{blue}{NTPP} \\
\hline
\textbf{NTPP} & \textcolor{blue}{NTPP} & \textcolor{green}{DC} & \textcolor{green}{EC} & \textcolor{green}{PO} & \textcolor{blue}{TPP} & \textcolor{green}{DC} & \textcolor{blue}{NTPP} \\
\hline
\textbf{TPPi} & \textcolor{blue}{TPPi} & \textcolor{blue}{NTPPi} & \textcolor{green}{DC} & \textcolor{red}{EC} & \textcolor{green}{PO} & \textcolor{green}{NTPPi} & \textcolor{blue}{NTPPi} \\
\hline
\textbf{NTPPi} & \textcolor{blue}{NTPPi} & \textcolor{blue}{NTPPi} & \textcolor{green}{DC} & \textcolor{blue}{DC} & \textcolor{blue}{DC} & \textcolor{green}{DC} & \textcolor{blue}{EC} \\
\hline

\end{tabular}

%% file: table-xrcc8-composition-results.tex
\begin{tabular}{|c|c|c|c|c|c|c|c|}

\multicolumn{1}{c}{} \\ 
\hline
\textbf{claude-3-5-sonnet-20240620} & \textbf{DC} & \textbf{EC} & \textbf{PO} & \textbf{TPP} & \textbf{NTPP} & \textbf{TPPi} & \textbf{NTPPi} \\
\hline
\textbf{DC} & D,E,P,T,N,t,n,Q & D,E,\textcolor{blue}{+P},\textcolor{blue}{+T},\textcolor{blue}{+N} & D,E,P,\textcolor{blue}{+T},\textcolor{blue}{+N} & D,E,\textcolor{blue}{+P},\textcolor{blue}{+T},\textcolor{blue}{+N} & D,\textcolor{blue}{+E},\textcolor{blue}{+P},\textcolor{blue}{+T},\textcolor{blue}{+N} & D & D \\
\hline
\textbf{EC} & D,E,\textcolor{blue}{+P},\textcolor{blue}{+t},\textcolor{blue}{+n} & D,E,P,\textcolor{blue}{+T},\textcolor{blue}{+t},\textcolor{blue}{+Q} & D,E,P,\textcolor{blue}{+T},\textcolor{blue}{+N} & \textcolor{red}{-D},E,P,T,\textcolor{blue}{+N} & \textcolor{red}{-D},\textcolor{red}{-E},\textcolor{blue}{+P},\textcolor{blue}{+T},\textcolor{blue}{+N} & D,E,\textcolor{red}{-P} & D \\
\hline
\textbf{PO} & D,E,\textcolor{blue}{+P},\textcolor{blue}{+t},\textcolor{blue}{+n} & D,E,P,\textcolor{blue}{+t},\textcolor{blue}{+n} & D,E,P,\textcolor{blue}{+T},\textcolor{blue}{+N},\textcolor{blue}{+t},\textcolor{blue}{+n},\textcolor{blue}{+Q} & \textcolor{red}{-D},\textcolor{red}{-E},P,T,N & \textcolor{red}{-D},\textcolor{red}{-E},P,T,N & \textcolor{blue}{+D},\textcolor{blue}{+E},P,\textcolor{red}{-T},\textcolor{red}{-N},\textcolor{blue}{+t},\textcolor{blue}{+n},\textcolor{red}{-Q} & D,E,P,\textcolor{red}{-T},\textcolor{red}{-N},\textcolor{blue}{+t},\textcolor{blue}{+n} \\
\hline
\textbf{TPP} & D & D,E,\textcolor{red}{-T} & D,E,P,T,N & T,N & \textcolor{red}{-T},N & \textcolor{blue}{+D},E,P,T,\textcolor{red}{-N},t,\textcolor{red}{-n},Q & D,E,P,\textcolor{red}{-T},\textcolor{red}{-N},\textcolor{blue}{+t},\textcolor{blue}{+n} \\
\hline
\textbf{NTPP} & D & D & D,E,P,T,N & \textcolor{red}{-T},N & \textcolor{red}{-T},N & \textcolor{blue}{+D},\textcolor{blue}{+E},\textcolor{blue}{+P},T,N & \textcolor{blue}{+D},E,P,T,N,t,n,Q \\
\hline
\textbf{TPPi} & D,\textcolor{blue}{+E},\textcolor{blue}{+P},\textcolor{blue}{+t},\textcolor{blue}{+n} & \textcolor{red}{-D},E,P,t,\textcolor{blue}{+n} & P,\textcolor{blue}{+t},\textcolor{blue}{+n} & P,T,t,Q & P,T,N & t,n & \textcolor{red}{-t},n \\
\hline
\textbf{NTPPi} & D,\textcolor{blue}{+E},\textcolor{blue}{+P},\textcolor{blue}{+t},\textcolor{blue}{+n} & \textcolor{red}{-D},\textcolor{red}{-E},P,\textcolor{blue}{+t},\textcolor{blue}{+n} & P,t,n & \textcolor{blue}{+P},t,n & P,T,N,t,n,Q & \textcolor{red}{-t},n & n \\
\hline

\multicolumn{1}{c}{} \\ 
\hline
\textbf{azure-gpt-35-turbo-0125} & \textbf{DC} & \textbf{EC} & \textbf{PO} & \textbf{TPP} & \textbf{NTPP} & \textbf{TPPi} & \textbf{NTPPi} \\
\hline
\textbf{DC} & D,\textcolor{blue}{+E},\textcolor{blue}{+P},\textcolor{blue}{+T},\textcolor{blue}{+N},\textcolor{blue}{+t},\textcolor{blue}{+n},\textcolor{blue}{+Q} & \textcolor{blue}{+D},\textcolor{blue}{+E},P,T,N,\textcolor{red}{-t},\textcolor{red}{-n},\textcolor{red}{-Q} & \textcolor{blue}{+D},\textcolor{blue}{+E},\textcolor{blue}{+P},\textcolor{blue}{+T},\textcolor{blue}{+N} & \textcolor{blue}{+D},\textcolor{blue}{+E},\textcolor{blue}{+P},\textcolor{blue}{+T},N,\textcolor{red}{-Q} & \textcolor{blue}{+D},\textcolor{blue}{+E},\textcolor{blue}{+P},T,N & \textcolor{blue}{+D},\textcolor{red}{-T},\textcolor{red}{-N} & \textcolor{blue}{+D},\textcolor{red}{-N},\textcolor{red}{-Q} \\
\hline
\textbf{EC} & D,E,\textcolor{blue}{+P},\textcolor{blue}{+t},\textcolor{blue}{+n} & D,E,\textcolor{blue}{+P},\textcolor{blue}{+T},\textcolor{blue}{+t},\textcolor{blue}{+Q} & \textcolor{blue}{+D},\textcolor{blue}{+E},\textcolor{blue}{+P},T,N & \textcolor{blue}{+E},\textcolor{blue}{+P},\textcolor{blue}{+T},N,\textcolor{red}{-t} & P,T,\textcolor{blue}{+N} & \textcolor{blue}{+D},\textcolor{blue}{+E},\textcolor{red}{-T},\textcolor{red}{-N} & \textcolor{blue}{+D},\textcolor{red}{-T},\textcolor{red}{-N} \\
\hline
\textbf{PO} & D,\textcolor{blue}{+E},P,\textcolor{blue}{+t},\textcolor{blue}{+n} & \textcolor{blue}{+D},\textcolor{blue}{+E},\textcolor{blue}{+P},\textcolor{red}{-N},\textcolor{blue}{+t},n & D,\textcolor{blue}{+E},P,\textcolor{blue}{+T},\textcolor{blue}{+N},\textcolor{blue}{+t},\textcolor{blue}{+n},\textcolor{blue}{+Q} & \textcolor{blue}{+P},\textcolor{blue}{+T},N,\textcolor{red}{-n} & \textcolor{blue}{+P},T,N & \textcolor{blue}{+D},\textcolor{blue}{+E},\textcolor{blue}{+P},\textcolor{red}{-N},\textcolor{blue}{+t},n & \textcolor{blue}{+D},\textcolor{blue}{+E},\textcolor{blue}{+P},\textcolor{red}{-T},\textcolor{red}{-N},\textcolor{blue}{+t},\textcolor{blue}{+n} \\
\hline
\textbf{TPP} & D,\textcolor{red}{-N},\textcolor{red}{-n} & \textcolor{blue}{+D},\textcolor{blue}{+E},\textcolor{red}{-N},\textcolor{red}{-t} & D,\textcolor{blue}{+E},\textcolor{blue}{+P},\textcolor{blue}{+T},N & T,N & \textcolor{blue}{+N},\textcolor{red}{-t},\textcolor{red}{-n} & \textcolor{blue}{+D},\textcolor{blue}{+E},\textcolor{blue}{+P},\textcolor{blue}{+T},\textcolor{red}{-N},\textcolor{blue}{+t},\textcolor{red}{-n},\textcolor{blue}{+Q} & \textcolor{blue}{+D},\textcolor{blue}{+E},\textcolor{blue}{+P},\textcolor{red}{-N},t,\textcolor{blue}{+n} \\
\hline
\textbf{NTPP} & D,\textcolor{red}{-N} & \textcolor{blue}{+D},\textcolor{red}{-E},\textcolor{red}{-N} & \textcolor{blue}{+D},\textcolor{blue}{+E},P,\textcolor{blue}{+T},N & \textcolor{red}{-T},N,\textcolor{red}{-t} & N,\textcolor{red}{-Q} & \textcolor{blue}{+D},\textcolor{blue}{+E},\textcolor{blue}{+P},T,N & \textcolor{blue}{+D},\textcolor{blue}{+E},\textcolor{blue}{+P},\textcolor{blue}{+T},N,\textcolor{blue}{+t},n,\textcolor{blue}{+Q} \\
\hline
\textbf{TPPi} & \textcolor{blue}{+D},\textcolor{blue}{+E},\textcolor{blue}{+P},\textcolor{red}{-N},\textcolor{blue}{+t},n & \textcolor{red}{-D},E,P,\textcolor{red}{-T},\textcolor{red}{-N},\textcolor{blue}{+t},\textcolor{blue}{+n},\textcolor{red}{-Q} & \textcolor{blue}{+P},\textcolor{red}{-T},\textcolor{red}{-N},\textcolor{blue}{+t},\textcolor{blue}{+n} & \textcolor{blue}{+P},\textcolor{blue}{+T},\textcolor{red}{-N},\textcolor{blue}{+t},\textcolor{red}{-n},\textcolor{blue}{+Q} & \textcolor{blue}{+P},T,N & \textcolor{red}{-T},\textcolor{red}{-N},\textcolor{blue}{+t},\textcolor{blue}{+n} & \textcolor{red}{-T},\textcolor{red}{-N},\textcolor{blue}{+n} \\
\hline
\textbf{NTPPi} & D,\textcolor{blue}{+E},\textcolor{blue}{+P},\textcolor{red}{-N},\textcolor{blue}{+t},\textcolor{blue}{+n} & \textcolor{red}{-D},\textcolor{blue}{+P},\textcolor{red}{-N},\textcolor{blue}{+t},\textcolor{blue}{+n} & \textcolor{red}{-D},\textcolor{blue}{+P},\textcolor{red}{-N},\textcolor{blue}{+t},\textcolor{blue}{+n} & \textcolor{blue}{+P},\textcolor{red}{-N},t,\textcolor{blue}{+n} & \textcolor{blue}{+P},T,N,\textcolor{blue}{+t},\textcolor{blue}{+n},\textcolor{blue}{+Q} & \textcolor{red}{-T},\textcolor{red}{-N},\textcolor{blue}{+n} & \textcolor{red}{-N},\textcolor{blue}{+n},\textcolor{red}{-Q} \\
\hline

\end{tabular}

%% file: table-xrcc8-preferred-results.tex
\begin{tabular}{|c|c|c|c|c|c|c|c|}

\hline
\textbf{claude-3-5-sonnet-20240620} & \textbf{DC} & \textbf{EC} & \textbf{PO} & \textbf{TPP} & \textbf{NTPP} & \textbf{TPPi} & \textbf{NTPPi} \\
\hline
\textbf{DC} & \textcolor{green}{DC} & \textcolor{green}{DC} & \textcolor{green}{EC} & \textcolor{red}{EC} & \textcolor{blue}{EC} &  & \textcolor{green}{DC} \\
\hline
\textbf{EC} & \textcolor{green}{DC} & \textcolor{blue}{PO} & \textcolor{green}{PO} & \textcolor{green}{PO} & \textcolor{green}{DC} & \textcolor{blue}{PO} & \textcolor{blue}{PO} \\
\hline
\textbf{PO} &  & \textcolor{blue}{EC} & \textcolor{blue}{PO} & \textcolor{blue}{TPP} &  & \textcolor{blue}{PO} & \textcolor{red}{NTPP} \\
\hline
\textbf{TPP} &  & \textcolor{green}{DC} &  & \textcolor{blue}{PO} &  &  & \textcolor{blue}{NTPP} \\
\hline
\textbf{NTPP} & \textcolor{green}{PO} & \textcolor{green}{DC} & \textcolor{green}{EC} & \textcolor{green}{PO} & \textcolor{blue}{PO} & \textcolor{green}{DC} & \textcolor{green}{PO} \\
\hline
\textbf{TPPi} & \textcolor{blue}{TPPi} &  & \textcolor{green}{DC} & \textcolor{red}{EC} & \textcolor{green}{PO} & \textcolor{blue}{PO} & \textcolor{blue}{PO} \\
\hline
\textbf{NTPPi} &  &  & \textcolor{green}{DC} &  &  & \textcolor{green}{DC} & \textcolor{green}{DC} \\
\hline

\multicolumn{1}{c}{} \\ 

\hline
\textbf{azureai-llama-3-70b-instruct} & \textbf{DC} & \textbf{EC} & \textbf{PO} & \textbf{TPP} & \textbf{NTPP} & \textbf{TPPi} & \textbf{NTPPi} \\
\hline
\textbf{DC} & \textcolor{green}{DC} & \textcolor{blue}{EC} & \textcolor{green}{EC} & \textcolor{blue}{NTPP} & \textcolor{blue}{EC} &  & \textcolor{green}{DC} \\
\hline
\textbf{EC} & \textcolor{blue}{EC} & \textcolor{blue}{PO} & \textcolor{green}{PO} & \textcolor{green}{PO} & \textcolor{green}{DC} & \textcolor{red}{TPP} & \textcolor{blue}{PO} \\
\hline
\textbf{PO} &  & \textcolor{blue}{EC} & \textcolor{blue}{PO} & \textcolor{blue}{TPP} &  & \textcolor{blue}{TPP} & \textcolor{red}{TPP} \\
\hline
\textbf{TPP} &  & \textcolor{green}{DC} &  & \textcolor{blue}{PO} &  &  & \textcolor{blue}{NTPP} \\
\hline
\textbf{NTPP} & \textcolor{blue}{NTPP} & \textcolor{green}{DC} & \textcolor{green}{EC} & \textcolor{green}{PO} & \textcolor{blue}{TPP} & \textcolor{green}{DC} & \textcolor{blue}{NTPP} \\
\hline
\textbf{TPPi} & \textcolor{blue}{TPPi} &  & \textcolor{green}{DC} & \textcolor{red}{EC} & \textcolor{green}{PO} & \textcolor{red}{TPP} & \textcolor{blue}{NTPP} \\
\hline
\textbf{NTPPi} &  &  & \textcolor{green}{DC} &  &  & \textcolor{green}{DC} & \textcolor{blue}{EC} \\
\hline

\end{tabular}